\documentclass[a4paper]{cas-sc}

\usepackage[authoryear]{natbib}

\usepackage{bm}
\usepackage{amssymb}
\usepackage{amsmath}
\usepackage{hyperref}
\usepackage{graphicx}
\usepackage{multirow}
\usepackage{placeins}
\usepackage{cleveref}
\usepackage{floatflt}
\crefname{figure}{Fig.}{Fig.}
\crefname{table}{Table}{Table}
\crefname{section}{}{}
\usepackage[normalem]{ulem}
\usepackage{caption}
\useunder{\uline}{\ul}{}
\def\tsc#1{\csdef{#1}{\textsc{\lowercase{#1}}\xspace}}
\tsc{WGM}
\tsc{QE}
\tsc{EP}
\tsc{PMS}
\tsc{BEC}
\tsc{DE}
\makeatletter
\def\@author#1{\g@addto@macro\elsauthors{\normalsize%
		\def\baselinestretch{1}%
		\upshape\authorsep#1\unskip\textsuperscript{%
			\ifx\@fnmark\@empty\else\unskip\sep\@fnmark\let\sep=,\fi
			\ifx\@corref\@empty\else\unskip\sep\@corref\let\sep=,\fi
		}%
		\def\authorsep{\unskip,\space}%
		\global\let\@fnmark\@empty
		\global\let\@corref\@empty  %% Added
		\global\let\sep\@empty}%
	\@eadauthor={#1}
}
\makeatother

\begin{document}
	\let\WriteBookmarks\relax
	\def\floatpagepagefraction{1}
	\def\textpagefraction{.001}
	\shorttitle{Paying more attention to local contrast: improving infrared small target detection performance via prior knowledge}
	\shortauthors{Peichao Wang et~al.}
	
	\title [mode = title]{Paying more attention to local contrast: improving infrared small target detection performance via prior knowledge}                      
	
	\author{Peichao Wang}
	\affiliation{organization={Command and Control Engineering College},addressline={Army Engineering University of PLA}, 
		city={Nanjing},
		postcode={210017}, 
		state={Jiangsu},
		country={China}}
	
	\author{Jiabao Wang}
	\author{Yao Chen}
	\author{Rui Zhang\corref{corresponding}}
	\cormark[1]
	\ead{rzhang_aeu@163.com}
	\author{Yang Li}
	\author{Zhuang Miao}
	\cortext[cor1]{Corresponding author}
	\begin{abstract}
The data-driven method for infrared small target detection (IRSTD) has achieved promising results. However, due to the small scale of infrared small target datasets and the limited number of pixels occupied by the targets themselves, it is a challenging task for deep learning methods to directly learn from these samples. Utilizing human expert knowledge to assist deep learning methods in better learning is worthy of exploration. To effectively guide the model to focus on targets' spatial features, this paper proposes the Local Contrast Attention Enhanced infrared small target detection Network (LCAE-Net), combining prior knowledge with data-driven deep learning methods. LCAE-Net is a U-shaped neural network model which consists of two developed modules: a Local Contrast Enhancement (LCE) module and a Channel Attention Enhancement (CAE) module. The LCE module takes advantages of prior knowledge, leveraging handcrafted convolution operator to acquire Local Contrast Attention (LCA), which could realize background suppression while enhance the potential target region, thus guiding the neural network to pay more attention to potential infrared small targets' location information. To effectively utilize the response information throughout downsampling progresses, the CAE module is proposed to achieve the information fusion among feature maps' different channels. Experimental results indicate that our LCAE-Net outperforms existing state-of-the-art methods on the three public datasets NUDT-SIRST, NUAA-SIRST, and IRSTD-1K, and its detection speed could reach up to 70 fps. Meanwhile, our model has a parameter count and Floating-Point Operations (FLOPs) of 1.945M and 4.862G respectively, which is suitable for deployment on edge devices.
	\end{abstract}
	
	\let\printorcid\relax
	\maketitle
\section{Introduction}
\par Visible light imaging is easily disrupted by external environmental factors and can be difficult to operate effectively in complex environments. Comparing to it, infrared imaging has obvious advantages in anti-interference \citep{ref0}. Additionally, infrared imaging can operate at long distances in all-weather, making it widely used in fields such as early warning and space-based surveillance systems \citep{ref1}. Compared to radar, infrared detection equipment does not need to actively emit detection signals and possess strong concealment, thus playing a significant role in complex battlefield environments. Effectively detecting of the infrared small targets has always been an important and challenging task in this field. According to the Society of Photo-Optical Instrumentation Engineers (SPIE), typical infrared small targets have characteristics such as contrast ratio less than 15\%, Signal-to-Noise Ratio (SNR) less than 1.5, and target size less than 0.15\% of the entire image \citep{ref2}.
\par To effectively detect infrared small targets, researchers have proposed a large number of methods. Existing methods can be broadly divided into two categories: model-driven methods and data-driven methods \citep{ref3}. Model-driven methods primarily involve handcrafted models and do not require the support of large amounts of data. These methods usually follow specific assumptions, with the aim of purely using experts' prior knowledge to find out the targets, such as looking for outliers with discontinuous grayscale in slowly transitioning backgrounds \citep{ref4}. Model-driven methods are significantly impacted by backgrounds and noise, exhibiting heightened sensitivity to hyperparameters such as pre-defined window sizes and segmentation thresholds. This necessitates iterative fine-tuning and experimentation tailored to specific scenarios, ultimately compromising their robustness and adaptability. Data-driven methods mainly rely on deep learning models, leveraging supervised learning techniques to empower the deep neural networks with the capability to autonomously learn features from diverse datasets, demonstrating robust generalization across varied scenes. With the rapid advancements of deep learning technologies, infrared small target detection (IRSTD) methods based on Convolutional Neural Networks (CNN) \citep{ref5,ref46,ref47} and transformer architectures \citep{ref6,ref48,ref49} have been proposed, which have achieved good detection results. However, the challenges associated with acquiring high-quality infrared images and annotations significantly constrain purely data-driven methods. Additionally, CNN and transformer-based networks typically encompass a vast number of parameters, posing difficulties in their deployment on edge devices with limited computational capabilities.
\par In recent years, hybrid detection methods that combine model-driven and data-driven approaches have brought new possibilities. By seamlessly integrating  prior knowledge inherent in model-driven methods into deep learning frameworks, these hybrid methods have witnessed improvements in terms of model performance, parameter size and computational complexity. Researchers have proposed hybrid detection models including RDIAN (Receptive-Field and Direction Induced Attention Network) \citep{ref5}, ALCNet (Attentional Local Contrast Network) \citep{ref7} and MSDA-Net (Multi-Scale Direction-Aware Network) \citep{ref8}, and so on. These methods mainly insight from human visual system (HVS) and create multiscale dilated convolution operators to emphasis one pixel's local contrast. However, different scale may pose different effects on detection performance, it is wise to figure out the influence of one operator instead of directly adding or concatenating multiscale operators to create new feature maps. Based on this intuition, we propose the Local Contrast Enhancement (LCE) module. In this module, we compute the Local Contrast Attention (LCA) as an indicator to emphasize potential targets while suppressing the background surrounding them, guiding neural networks to precisely pinpoint potential locations of infrared small targets. Furthermore, to facilitate efficient cross-channel information fusion and maintain semantic information of small targets during downsampling stages, we present the Channel Attention Enhancement (CAE) module. Combining these two innovative modules, we propose the Local Contrast Attention Enhanced infrared small target detection Network (LCAE-Net). This model has advantages in parameter size and Floating-Point Operations (FLOPs) compared to current detection methods and has achieved excellent test results on the three benchmark datasets\text{—}NUAA-SIRST, NUDT-SIRST, and IRSTD-1K, effectively striking a balance between performance and computational efficiency.
\par In summary, our contributions are summarized below:
\begin{itemize}
	\item[$\bullet$]LCAE-Net is proposed as the integration of model-driven and data-driven method, which introduces prior knowledge to sufficiently utilize local contrast information.  It is an effective U-shaped neural network model that fully utilizes prior knowledge to boost the model performance. 
	\item[$\bullet$]A LCE module which leverages prior knowledge to enhance the local contrast features of potential infrared small targets is proposed, effectively guiding the constructed neural network model to capture the location information of potential targets. 
	\item[$\bullet$]A CAE module is proposed, achieving fusion of information among different channels in the multi-scale feature maps obtained from downsampling.
	\item[$\bullet$]Experiments on three public benchmark datasets demonstrated remarkable performance of our method. Comparing to several state-of-the-art detection methods, LCAE-Net has successfully achieved a balance between superior performance and efficient computation, indicating its potential effectiveness on edge devices.
\end{itemize}
\section{Related works}
\subsection{Model-driven infrared small target detection methods}
\par In early stages, IRSTD tasks mainly relied on traditional model-based methods, which can be further divided into background suppression-based methods, optimization-based methods, and HVS-based methods. Background suppression-based methods \citep{ref10,ref9,ref14} regard infrared small targets as outliers in the original image and obtain target information through filtering algorithms, while optimization-based methods  \citep{ref15,ref16,ref21} consider infrared small targets have sparse characteristics in images which could be separated from background pixels through matrix decomposition. Comparing these two kinds of methods, HVS-based methods have more physical meanings. 
\par Contrast is the most important quantity in the visual system’s flow coding \citep{ref22}, and the human visual system can be quickly attracted to small targets on infrared images \citep{ref23}. Inspired by this, methods based on HVS learn from selective mechanism of the human eye’s rapid response to visually salient regions, using local contrast to achieve the detection of infrared small targets. \citet{ref22} proposed the classic Local Contrast Measure (LCM) method, which uses local contrast features to highlight the target and employs adaptive thresholding for target extraction. \citet{ref24} improved the LCM’s limitation to bright targets and proposed the MPCM (Multiscale Patch-based Contrast Measure) method, which enhances both bright and dark targets and uses parallel computing to improve real-time performance. \citet{ref25} further proposed the Local Energy Factor (LEF) based on LCM, an indicator for measuring the dissimilarity between the target area and the surrounding background, enriching the description of local image differences, and achieving target segmentation and extraction through adaptive thresholding when combined with LCM. \citet{ref26} proposed the variance difference weighted three-layer local contrast measure (VDWTLCM) method to effectively suppress highlight backgrounds and strong noise, where the three-layer local contrast measure (TLCM) module calculates the local contrast of a single pixel for target enhancement and background suppression, and the weighting function based on the mean value of the variance difference (MVD) further suppresses prominent background edges. \citet{ref27} used density peak search to find candidate target positions in images preprocessed with Gaussian differentiation, enhanced the gradient saliency features of candidate targets using local contrast methods for background suppression, and determined the target position through threshold segmentation after calculating and fusing multi-directional gradient features. Contrast-based algorithms are designed by referencing the principles of how the human visual system observes objects and have relatively strong interpretability, providing the possibility of integrating prior knowledge into deep learning models.
\subsection{Data-driven infrared small target detection methods}
\par Due to the limited number of pixels occupied by infrared small targets in images, directly applying typical object detection methods such as R-CNN \citep{ref53}, YOLO \citep{ref54}, and SSD \citep{ref55} to IRSTD field often leads to poor performance. As a result, researchers usually model it as a pixel-level binary classification problem \citep{ref28}.
\par IRSTD methods based on CNN are currently prevalent in research. \citet{ref29} introduced the Asymmetric Contextual Modulation (ACM) module, which effectively realized the fusion of shallow and deep features, with its efficacy validated through experiments on Feature Pyramid Network (FPN) and U-Net. \citet{ref30} relieved the issue of response vanishing due to deeper network layers by proposing the DNA-Net (Dense Nested Attention Network), which repeatedly fuses and enhances multi-level features through dense connections, fully integrating and utilizing the contextual information of small targets. \citet{ref31} proposed the ISTDU-Net (Infrared Small-Target Detection U-Net), which improves the downsampling and skip connections of traditional U-Net by introducing feature map groups and fully connected layers to enhance small target weights and increase the global receptive field. \citet{ref32} proposed the ISNet (Infrared Shape Network), exploring the use of the Taylor Finite Difference (TFD) module for edge detection and the Two-Orientation Attention Aggregation (TOAA) module for the fusion of low-level and high-level information in both row and column directions, capturing target shape features and suppressing noise. \citet{ref33} developed the UIU-Net (U-Net in U-Net), embedding a small U-Net within a large U-Net backbone network to achieve multi-level and multi-scale representation learning of targets while mitigating the issue of deep response vanishing for small targets. \citet{ref34} pointed out that while much research focuses on the design of the network itself, there is insufficient exploration of the loss function area, thus they proposed a new Scale and Location Sensitive (SLS) loss to assist U-shaped networks in distinguishing and locating targets of different scales, and further developed the MSHNet (Multi-Scale Head to the plain U-Net) based on this loss function. With the introduction of Vision Transformer (ViT), the application of Transformer models in IRSTD tasks \citep{ref35,ref36,ref6} has become increasingly common. Additionally, methods based on diffusion models \citep{ref37}, graph neural networks \citep{ref38} and state space models \citep{ref56} have also been proposed. Purely data-driven methods often resort to complex connections among different layers to preserve small targets' responses, and promote model performance at the cost of increased parameters and computational costs.
\subsection{Hybrid infrared small target detection methods}
\par Improving model performance at the cost of increasing model complexity is not a long-term solution. Since model-driven methods effectively utilize the expertise of human experts, combining them with deep learning methods can effectively integrate prior knowledge into deep learning models, thereby enhancing the detection performance of proposed models. 
\par Insights in HVS are popular in this field. \citet{ref7} drew inspiration from DeepLab series \citep{ref44} and MPCM \citep{ref24}, embedding traditional multi-scale local contrast method into the end-to-end network model ALCNet as a parameter-free linear layer, which realized a integration of model-driven and data-driven approaches. \citet{ref39} proposed a Multiscale Local Contrast Learning Network (MLCL-Net) that integrates local contrast into network training to generate local contrast feature maps. \citet{ref5} considered that a small target usually appears as a bright spot area with a Gaussian-like distribution, and grayscale differences between the central and boundary pixels exist in all the directions, which is known as multidirectionality. Inspired by this prior knowledge, they proposed RDIAN, which consists of one Multi-Directional Target Enhancement (MDTE) module aiming at enhancing target features in low-level feature maps. \citet{ref45} changed the approach of acquiring local contrast and developed multiple attention local contrast module stemmed from ALCNet, then integrated it into their developed Local Contrast Attention Guide Network (LCAGNet) to promote detection performance. \citet{ref8} utilized the prior knowledge of targets and constructed Multi-Directional Feature Awareness (MDFA) module to emphasize the focus on high-frequency directional features. However, due to the small area occupied by infrared small targets in the image, applying multiscale convolution operators to obtain feature maps at different scales may not necessary; in fact, even with single-scale convolution operator, we can achieve good results with lower computational complexity guided by prior knowledge.
\section{Proposed method}
\subsection{Overall pipeline}
\par
\begin{figure*}[htp]\centering
	\includegraphics[width=\linewidth]{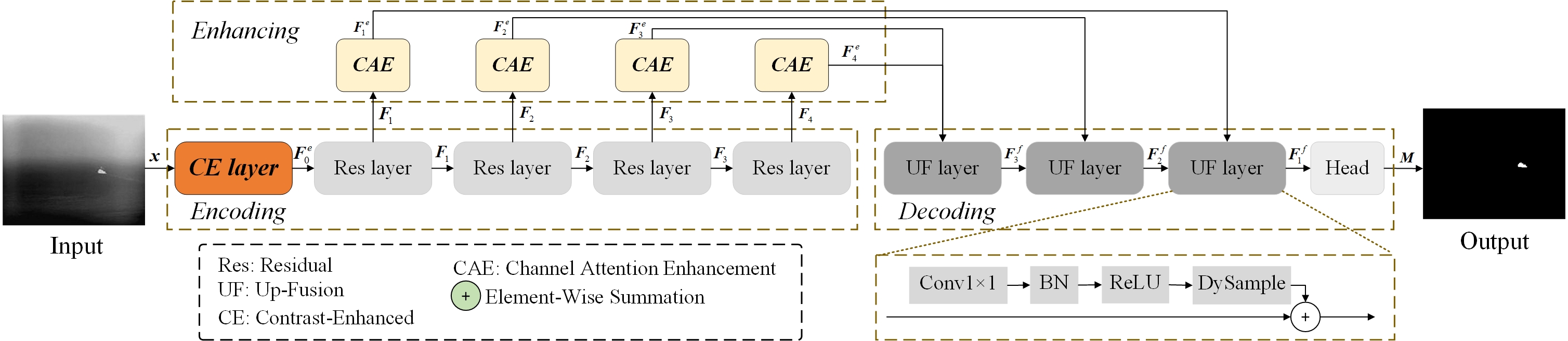}
	\caption{Overview of the proposed LCAE-Net.}
	\label{fig_1}
	\vspace{-\baselineskip}
\end{figure*}

 \cref{fig_1} illustrates the main framework of LCAE-Net proposed in this paper. The framework comprises three primary stages: an encoding stage, an enhancing stage, and a decoding stage. 
\par Given an infrared grayscale image $\bm{x} \in {\bm{R}^{1 \times H \times W}}$, where $H$ and $W$ denote the height and width of the image. LCAE-Net initially employs the encoding stage to perform channel expansion and acquire multi-scale feature maps. The residual layer expands the channels of the input feature map $\bm{F}_0^e$ and performs spatial downsampling, thereby extracting a multi-scale feature map that contains contextual semantic information. Each layer consists of different numbers of ResNeSt blocks, denoted by  ${N_i}(i = 1,2,3,4)$ which represents the number of blocks in each layer. For the input feature map $\bm{F}_0^e$, after being processed through each layer, the resulting feature map  ${\bm{F}_i} \in {\bm{R}^{{C_i} \times {H_i} \times {W_i}}}$ is obtained, where ${C_i}$, ${H_i}$ and ${W_i}$ represent the number of channels, height, and width of the obtained feature map respectively. The ResNeSt block is the main building block of the ResNeSt network \citep{ref40}, which draws inspiration from SENet \citep{ref41} and ResNeXt \citep{ref42}. It introduces the concept of a cardinal group, where each group uses split-attention (SA) to obtain weights between channels, enhancing the interaction between different channels of information and improving the performance of downstream tasks. In our paper, we learn from the design concept of ISTDU-Net and sets ${N_i}$ to 1, 2, 4, and 8. Through the implementation of SA, the correlation between channels of the input feature map is effectively enhanced, thereby improving the ability of cross-channel information interaction. This enables the model to better achieve the extraction of high-quality feature maps.
\par Subsequently, the enhancing stage effectively integrates and enhances information among different channels of four residual layers' output feature maps, preventing feature degradation from direct summation, and forwards the enhanced feature maps to the decoding stage.
\par The decoding stage comprises three up-fusion layers and one prediction head. The up-fusion layer performs upsampling on the input low-level feature map, reducing the number of channels while increasing the spatial resolution. The low-level feature map $\bm{F}_4^e$, $\bm{F}_3^f$ and $\bm{F}_2^f$ with an input size of $2C \times {H \mathord{\left/
		{\vphantom {H 2}} \right.
		\kern-\nulldelimiterspace} 2} \times {W \mathord{\left/
		{\vphantom {W 2}} \right.
		\kern-\nulldelimiterspace} 2}$ undergoes a dimension reduction operation through a $1 \times 1$ convolution, reducing the dimension of the input feature map from $2C$ to $C$. Then the channel-reduced feature map is processed by a batch normalization layer and activated by ReLU function. We utilize the DySample \citep{ref43} lightweight dynamic upsampling operator on it to output a feature map of size $C \times H \times W$ that aligns with the feature map $\bm{F}_i^e(i = 1,2,3)$ of the same level. Subsequently, the two aligned feature maps are added together to realize feature fusion and reconstruction, outputting the fused feature map $\bm{F}_i^f$. After employing three consecutive feature fusion and reconstruction operations, the resulting feature map $\bm{F}_1^f$ is sent into the prediction head to generate point-by-point prediction results for each pixel. The prediction head processes the feature map $\bm{F}_1^f$ with $3 \times 3$ convolution, batch normalization layer, ReLU activation function and $1 \times 1$ convolution sequentially, and activates the output through the sigmoid function, thus a probability feature map ${\bm{M}'} \in {\bm{R}^{1 \times H \times W}}$ is generated. Each value on this feature map represents the probability of the corresponding pixel belonging to the target. Here, the value of the pixels with a probability greater than 0.5 is set to 1 while the rest are set to 0, resulting in a prediction mask $\bm{M} \in {\bm{R}^{1 \times H \times W}}$ which is consistent in size with the original input image.
\subsection{Local Contrast Enhancement module}
\par
\begin{figure*}[htp]\centering
	\includegraphics[width=\linewidth]{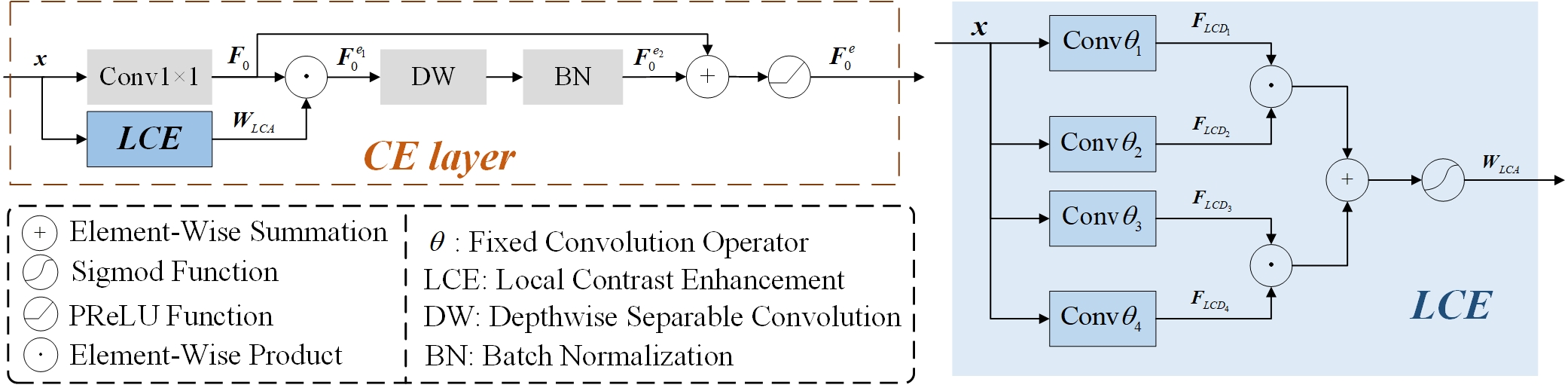}
	\caption{Structures of the CE layer and LCE module.}
	\label{fig_2}
	\vspace{-\baselineskip}
\end{figure*}

 The Local Contrast Enhancement (LCE) is the core module of the Contrast-Enhanced (CE) layer, which is the first layer of the proposed model. The CE layer processes the input infrared grayscale image, expanding its channels and enhancing extracted feature responses with the aid of LCE module. The LCE module is employed to derive attention weights for potential infrared small targets within the original image. These attention weights reflect probabilities whether current pixel is located in target areas. The LCE module's essence resides in fusing human expert knowledge into designing convolution operators. These operators integrate prior knowledge, effectively mitigating the blindness and tendency to fall into local extrema that often accompany with autonomous learning in deep neural networks. Specifically, as the first processing module, its optimization during training is highly susceptible to the influence of backpropagation gradients from subsequent modules. This can lead to issues such as slow parameter learning caused by gradient accumulation errors and a propensity to get stuck in local optima. By employing artificially designed convolution operators, we can effectively guide the neural network to focus on targets that we concern. The structures of the CE layer and LCE module are depicted in \cref{fig_2}.
\par
\begin{figure*}[htp]\centering
	\includegraphics[width=0.9\linewidth]{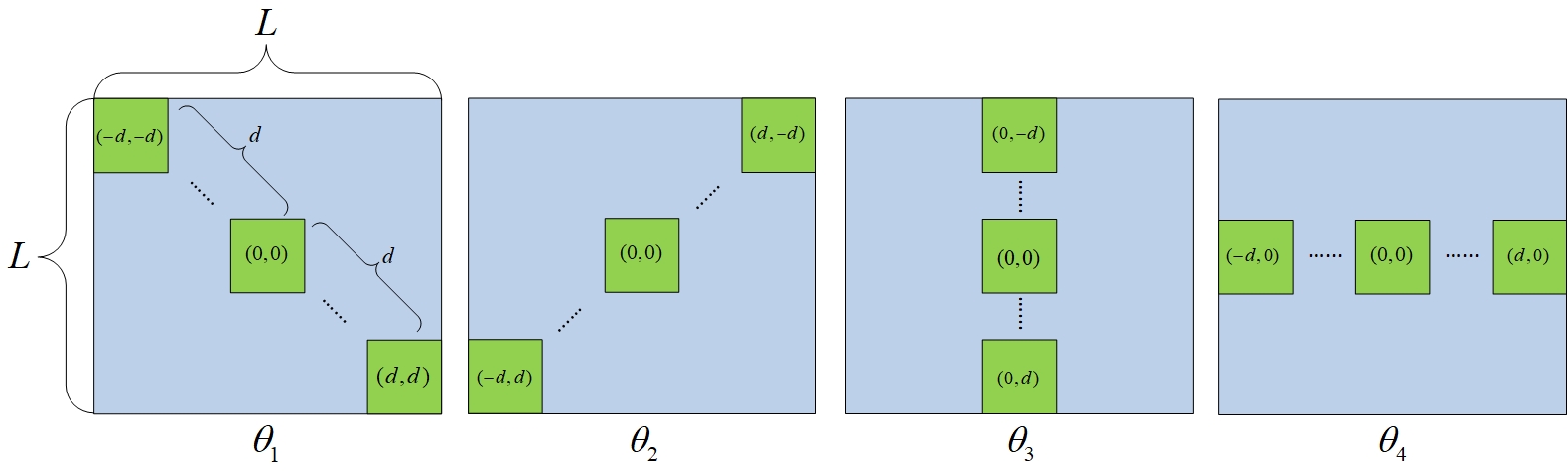}
	\caption{Structures of our fixed convolution operators.}
	\label{fig_3}
	\vspace{-\baselineskip}
\end{figure*}

 Given an infrared grayscale image $\bm{x} \in {\bm{R}^{1 \times H \times W}}$, it initially passes through a 1×1 convolution kernel for channel expansion,  yielding the primary feature map ${\bm{F}_0} \in {\bm{R}^{C \times H \times W}}$, where $C$ denotes the number of channels after expansion. Simultaneously, the original image is processed by the LCE module to output the matrix (attention weights) ${\bm{W}_{LCA}} \in {\bm{R}^{1 \times H \times W}}$. One infrared small target generally exhibits Gaussian-like distributions in images, and the gray value of small targets usually spreads uniformly in its surroundings \citep{ref5}. The gray value of an infrared small target is generally higher than surrounding background pixels, which means it has high local contrast. The visual attention mechanism is an important foundation for HVS, and human eyes will quickly focus on the important areas in the scene \citep{ref50}. Gaussian distribution typically features a bell-shaped curve, peaking in the middle and tapering off gradually towards the surrounding sides. Based on above prior knowledge, it is advisable for us to have a measure to present this local characteristic. Here, we propose the Local Contrast Distance ($LCD$) to precisely capture this local characteristic. For the input grayscale image $\bm{x}$, we define the following operations to acquire feature map $\bm{F}_{LC{D_i}}(i=1,2,3,4)$ which contains local contrast information:
\begin{equation}\label{eqn-1} 
	{\bm{F}_{LC{D_i}}} = {\theta _i} \otimes \bm{x}
\end{equation}
\par \noindent where $\otimes$ denotes convolution operation, $\theta_i$ denotes the $L \times L$ fixed convolution operator with dilation rate $d$, each value within $\bm{F}_{LC{D_i}}$ signifies the $LCD$ value of each pixel in the original image computed by using a fixed convolution operator $\theta_i$. Here, we illustrate the structure of each fixed convolution operator by setting its center coordinate as $(0,0)$, as depicted in \cref{fig_3}.
\par Each convolution operator, with the exception of green marked positions, comprises zeros in all other locations. Specifically, when applied to a pixel at coordinate $(m,n)$ within the original image $\bm{x}$, four $LCD$ values can be obtained through the following computations:
\begin{equation}\label{eqn-2} 
	\begin{aligned}
		&{\bm{F}_{LC{D_1}}}(m,n) = {\theta _1}(0,0) \times \bm{x}(m,n) - {\theta _1}(- d,- d) \times \bm{x}(m - d,n - d) - {\theta _1}(d,d) \times \bm{x}(m + d,n + d) \\
		&{\bm{F}_{LC{D_2}}}(m,n) = {\theta _2}(0,0) \times \bm{x}(m,n) - {\theta _2}(- d,d) \times \bm{x}(m - d,n + d) - {\theta _2}(d,- d) \times \bm{x}(m + d,n - d) \\ 
		&{\bm{F}_{LC{D_3}}}(m,n) = {\theta _3}(0,0) \times \bm{x}(m,n) - {\theta _3}(0,- d) \times \bm{x}(m,n - d) - {\theta _3}(0,d) \times \bm{x}(m,n + d) \\
		&{\bm{F}_{LC{D_4}}}(m,n) = {\theta _4}(0,0) \times \bm{x}(m,n) - {\theta _4}(- d,0) \times \bm{x}(m - d,n) - {\theta _4}(d,0) \times \bm{x}(m + d,n)\\
	\end{aligned}
\end{equation}
\par \noindent here $\bm{x}(m,n)$ denotes the gray scale of the pixel situated at the coordinate $(m,n)$ on the original image. Since infrared small targets exhibit a Gaussian-like distribution, where the central pixels within the target exhibit higher gray values while the boundary pixels possess relatively lower and comparable gray values, we set the central pixel value of fixed convolution operator as $\alpha$ and other non-zero values as $\beta$, thus the above formulas could be simplified as follows:
\begin{equation}\label{eqn-3} 
	\begin{aligned}
		&{\bm{F}_{LC{D_1}}}(m,n) = \alpha  \times \bm{x}(m,n) - \beta  \times (\bm{x}(m - d,n - d) + \bm{x}(m + d,n + d)) \\
		&{\bm{F}_{LC{D_2}}}(m,n) = \alpha  \times \bm{x}(m,n) - \beta  \times (\bm{x}(m,n - d) + \bm{x}(m,n + d))\\ 
		&{\bm{F}_{LC{D_3}}}(m,n) = \alpha  \times \bm{x}(m,n) - \beta  \times (\bm{x}(m - d,n + d) + \bm{x}(m + d,n - d)) \\
		&{\bm{F}_{LC{D_4}}}(m,n) = \alpha  \times \bm{x}(m,n) - \beta  \times (\bm{x}(m - d,n) + \bm{x}(m + d,n))\\
	\end{aligned}
\end{equation}
\par To further accentuate this difference, we define the Local Contrast Attention ($LCA$) weight for the pixel at the coordinate $(m,n)$ on the original image as follows:
\begin{equation}\label{eqn-4} 
	{\bm{W}_{LCA}}(m,n) = Sigmod ({\bm{F}_{LC{D_1}}}(m,n) \times {\bm{F}_{LC{D_2}}}(m,n) + {\bm{F}_{LC{D_3}}}(m,n) \times {\bm{F}_{LC{D_4}}}(m,n))
\end{equation}
\par
\begin{figure}[h]
	\centering
	\includegraphics[width=\linewidth]{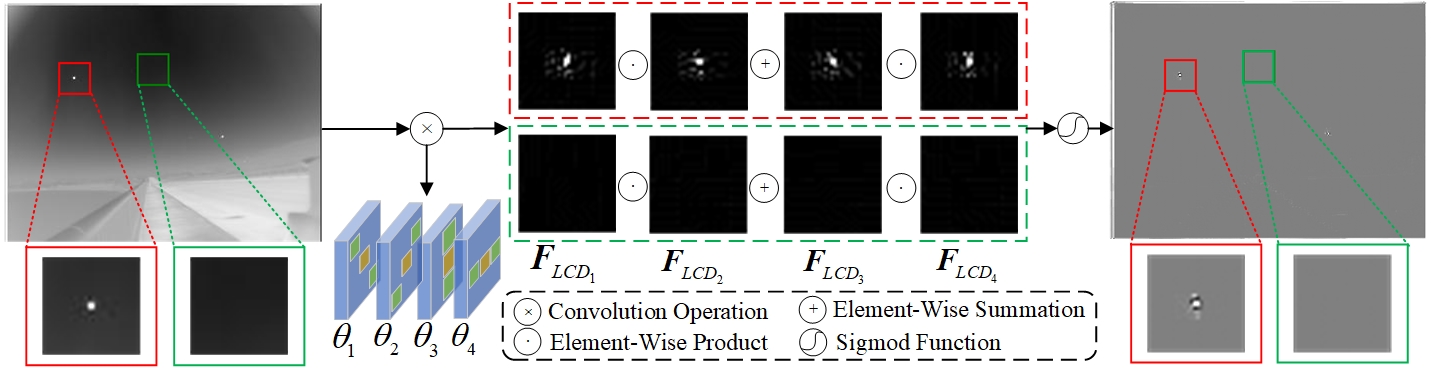}
	\caption{Illustration of the computation process of our LCE module. In our case, we set $\alpha$ as 1, $\beta$ as 0.5 and $d$ as 1.}
	\label{fig_4}
	\vspace{-\baselineskip}
\end{figure}

 The aforementioned procedure to acquire attention matrix $\bm{W}_{LCA}$ can be formally expressed using tensor operations:
\begin{equation}\label{eqn-5} 
	{\bm{W}_{LCA}} = {\bm{F}_{LC{D_1}}} \odot {\bm{F}_{LC{D_2}}} \oplus {\bm{F}_{LC{D_3}}} \odot {\bm{F}_{LC{D_4}}}
\end{equation}
\par \noindent here $\odot$ denotes the element-wise product, $\oplus$ denotes the element-wise summation. If current pixel lies in the target area, the calculated $LCD$ values should be larger, which means $LCA$ value would approximate 1; when the pixel is located within the background area, the calculated $LCD$ values should be smaller, which means $LCA$ value would approximate 0.5. We use a simple case here to illustrate above computations, which is depicted in \cref{fig_4}. In our case, we set $\alpha$ as 1, $\beta$ as 0.5 and $d$ as 1.
\FloatBarrier
\FloatBarrier
\par The red rectangular box in the image contains a small target, while the green rectangular box only contains background pixels. Upon examination of the visualization results, it becomes evident that within the attention matrix $\bm{W}_{LCA}$, the infrared small target stands out prominently in highlights, with the majority of remaining areas appearing predominantly gray. This indicates that irrelevant background information presenting in the original image has been effectively suppressed, highlighting the potential target areas. If we multiply $\bm{W}_{LCA}$ with the primary feature map ${\bm{F}_{0}}$, the pixels located in target area would have minimal change in gray level while those located in background area would have decrease in gray level. So the calculated attention matrix $\bm{W}_{LCA}$ are element-wise multiplied with the primary feature map ${\bm{F}_{0}}$ to obtain the enhanced feature map $\bm{F}_0^{{e_1}} \in {\bm{R}^{C \times H \times W}}$, which could guide the constructed neural network model to pay more attention to potential infrared small targets:
\begin{equation}\label{eqn-5} 
	\bm{F}_0^{{e_1}} = {\bm{W}_{LCA}} \odot {\bm{F}_0}
\end{equation}
\par After undergoing the processing of depthwise separable convolution and batch normalization, we obtain the resulting feature map $\bm{F}_0^{{e_2}} \in {\bm{R}^{C \times H \times W}}$, which mainly achieves the suppression of the background region. By element-wise adding $\bm{F}_0^{{e_2}}$ to ${\bm{F}_0}$, the useful contextual information of the primary feature map is preserved, completing background suppression while enhance the potential target region. Meanwhile, due to the introduction of residual structure, the proposed neural network model will be easier to optimize. It is followed by activation through the PReLU function to output the primary enhanced feature map $\bm{F}_0^e \in {\bm{R}^{C \times H \times W}}$. The computational process can be represented as follows:
\begin{equation}\label{eqn-7} 
	\begin{aligned}
		&\bm{F}_0^{{e_2}} = BN(DW(\bm{F}_0^{{e_1}})) \\
		&\bm{F}_0^e = PReLU({\bm{F}_0} + \bm{F}_0^{{e_2}})\\
	\end{aligned}
\end{equation}
\par \noindent where $DW$ denotes depthwise separable convolution and $BN$ denotes batch normalization layer.
\subsection{Channel Attention Enhancement module}
\begin{figure}[htp]\centering
	\includegraphics[width=0.9\linewidth]{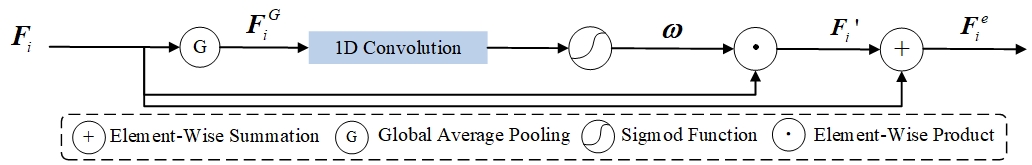}
	\caption{Structure of CAE module.}
	\label{fig_5}
	\vspace{-\baselineskip}
\end{figure}
\par With the progress of downsampling, the response information is distributed across the channels of output feature maps. To effectively utilize the information among different channels, fusing the information from each channel is a proven method. The CAE module is proposed to fuse information from different channels, and four CAE modules constitute the enhancing stage. By applying attention weighting to the multi-scale output feature map ${\bm{F}_i}(i = 1,2,3,4)$ from four residual layers in the encoding stage, the weighted feature map ${\bm{F}_i'}(i = 1,2,3,4)$ is obtained, which could reinforce the focus on potential infrared small target areas. By adding the original input feature map ${\bm{F}_i}$ to the weighted feature maps ${\bm{F}_i'}$ at the spatial level, the useful contextual information in the original input feature map is further fused with the channel-weighting information, ultimately generating the enhanced feature map $\bm{F}_i^e$. The structure of CAE module is illustrated in \cref{fig_5}.
\par For the input feature map ${\bm{F}_i} \in {\bm{R}^{{C_i} \times {H_i} \times {W_i}}}$, we initially apply global average pooling to obtain the feature representation $\bm{F}_i^G \in {\bm{R}^{1 \times {H_i} \times {W_i}}}(i = 1,2,3,4)$ based on global spatial information. Subsequently, inspired by \citep{ref52}, we use a one-dimensional convolution with a kernel size of 3 to interact with the current channel and its two adjacent channels, which could achieve the fusion of channel information with a relatively small number of parameters. The Sigmoid activation function is then used to output the weight values ${\omega _j} \in \bm{\omega} (j = 1,2,...,{\rm{C}})$ for each channel. These weight values are multiplied point-wise with the input feature map to obtain the weighted feature map ${\bm{F}_i'} \in {\bm{R}^{{C_i} \times {H_i} \times {W_i}}}$:
\begin{equation}\label{eqn-21} 
	\bm{F}_i' = {\bm{F}_i} \odot \bm{\omega}
\end{equation}
\par We then add this to original input feature map ${\bm{F}_i}$, ultimately yielding the enhanced feature map ${\bm{F}_i^e} \in {\bm{R}^{{C_i} \times {H_i} \times {W_i}}}(i = 1,2,3,4)$:
\begin{equation}\label{eqn-13} 
	\bm{F}_i^e = \bm{F}_i'  + {\bm{F}_i}
\end{equation}
\par By adding the channel-wise weighted result $\bm{F}_i'$ to the original input feature map $\bm{F}_i$ , the enhanced feature map could effectively retain the original contextual information while strengthening the semantic information at the locations of infrared small targets across different channel. This approach facilitates feature fusion and preservation of contextual information, aiding in the propagation of semantic information towards higher-resolution network layers during the decoding process, ultimately enabling precise pixel prediction. Furthermore, the integration of the weighted feature map with the original input feature map results in the adoption of a residual architecture within this module. This design facilitates the smooth backpropagation of gradients throughout the decoding and encoding stages, thereby enhancing the training efficiency of the LCAE-Net.
\section{Experiments}
\subsection{Experiments setting}
\subsubsection{Datasets}
\par The experiments conducted in this paper utilize three publicly available datasets for single-frame infrared small target detection: NUAA-SIRST \citep{ref29}, NUDT-SIRST \citep{ref30}, and IRSTD-1K \citep{ref32}. These datasets comprise 427, 1327 and 1000 images respectively. Here, we follow the rule from \citet{ref30} to partition the training and test sets of NUAA-SIRST and NUDT-SIRST, and from \citet{ref32} to split IRSTD-1K.
\subsubsection{Evaluation Metrics}
\par To verify the effectiveness of our method, we use several standard metrics.
\begin{enumerate}[(1)]
	\item $IoU$: It is defined as the ratio of the number of pixels in the intersection area between the predicted mask and the label mask to the number of pixels in the union area. The calculation formula is as follows:
	\begin{equation}\label{eqn-17} 
		I{\rm{o}}U = \frac{{{A_i}}}{{A{}_u}} = \frac{{\sum\nolimits_{i = 1}^N {T{P_i}} }}{{\sum\nolimits_{i = 1}^N {{T_i} + {P_i} - T{P_i}} }}
	\end{equation}
	Here ${A_i}$ and ${A_u}$ denote the number of pixels in the intersection region and union region respectively, $N$ denotes the number of samples tested, $T{P_i}$ denotes the number of accurately predicted pixels in the $i$-th sample, while ${T_i}$ and ${P_i}$ represent the number of pixels in the Ground Truth and the method’s prediction results for the sample to be tested respectively.
	\item ${P_d}$: ${P_d}$ calculates the ratio of the number of correctly detected targets ${N_{pred}}$ to the total number of targets ${N_{all}}$ in the image. The calculation formula is as follows:
	\begin{equation}\label{eqn-18}
		{P_d} = \frac{{{N_{pred}}}}{{{N_{all}}}}
	\end{equation}
	\par Here, following \citep{ref30}, we consider the target correctly predicted if the deviation of target centroid is less than 3.
	\item ${F_a}$: It calculates the ratio of the number of incorrectly predicted pixels ${N_{false}}$ to the total number of pixels ${P_{all}}$ in the image. The calculation formula is as follows:
	\begin{equation}\label{eqn-19}
		{F_a} = \frac{{{N_{false}}}}{{{P_{all}}}}
	\end{equation}
\end{enumerate}
\subsubsection{Implementation details}\label{Implementation details}
\par The experiments were conducted under the Ubuntu 18.04 operating system. The workstation utilized were equipped with one Intel i9-9900K central processing unit and 64 GB of memory, one NVIDIA GeForce RTX 3090 with 24 GB of graphics memory. All code was implemented using the Pytorch framework. During the training process, the input images were standardized and then resized to 256×256 pixels through edge padding and random cropping, and data augmentation was performed using random flipping. The number of training epochs was set to 400, the batch size was 16, and the loss function we used was the Soft-IoU loss \citep{ref52}. Parameters were updated using the Adam optimizer, with an initial learning rate of 0.0005, a momentum term of 0.9, and a polynomial decay learning rate scheduling strategy was adopted. The learning rate decay rate was set to 0.1, with adjustments made at epochs 200 and 300. In terms of hyperparameters, $\alpha$ was set to 1, $\beta $ was set to 0.5, and $d$ was set to 3. The values of other parameters and the results are detailed in Section \ref{Hyperparameter Analysis}.
\subsection{Results and analysis}
\par To evaluate the proposed method more objectively and comprehensively, we compare our model to the state-of-the-art (SOTA) IRSTD methods, including: DNA-Net \citep{ref30}, ALCNet \citep{ref7}, ACM \citep{ref29}, UIU-Net \citep{ref33}, RDIAN \citep{ref5}, ISTDU-Net \citep{ref31} and SCTransNet \citep{ref6}. Open-source implementations of these techniques can be found at github\footnote{\href{https://github.com/XinyiYing/BasicIRSTD}{https://github.com/XinyiYing/BasicIRSTD} and \href{https://github.com/xdFaiAll}{https://github.com/xdFaiAll}}. Comparative methods were trained in default hyperparameters from scratch using the same experimental settings. The performance of each method on the three public datasets is shown in \cref{table1}, where the optimal value and the suboptimal value in each column are highlighted in \textbf{bold} and \underline{underline} respectively.
\begin{table}[!h]
	\centering
	\caption{Comparisons with SOTA methods on three public datasets. The optimal value and the suboptimal value in each column are highlighted in \textbf{bold} and \underline{underline} respectively.}
	\label{table1}
	\begin{tabular}{cccccccccc}
		\hline
		\multirow{2}{*}{Method} & \multicolumn{3}{c}{NUAA-SIRST}                      & \multicolumn{3}{c}{NUDT-SIRST}                     & \multicolumn{3}{c}{IRSTD-1K}                        \\ \cline{2-10} 
		& $IoU$/\%         & ${P_d}$/\%           & ${F_a}$/${10^{ - 6}}$         & $IoU$/\%         & ${P_d}$/\%           & ${F_a}$/${10^{ - 6}}$        & $IoU$/\%         & ${P_d}$/\%           & ${F_a}$/${10^{ - 6}}$         \\ \hline
		DNA-Net \citep{ref30}                    & 74.092          & {\ul 95.420}    & 37.779          & 91.959          & 98.412          & 9.077          & {\ul 64.145}    & 89.562          & \textbf{14.670} \\
		ALCNet \citep{ref7}                     & 68.275          & 91.985          & 32.195          & 64.667          & 97.989          & 38.124         & 57.371          & 91.246          & 51.755          \\
		ACM \citep{ref29}                        & 69.729          & 90.840          & 31.229          & 66.576          & 96.402          & 19.464         & 56.915          & 91.582          & 82.709          \\
		UIU-Net \citep{ref33}                    & {\ul 76.950}    & 93.130          & {\ul 14.132}    & 87.231          & 97.883          & {\ul 1.907}    & 61.884          & 93.939          & 48.225          \\
		RDIAN \citep{ref5}                      & 71.558          & 95.038          & 52.394          & 81.928          & 97.566          & 10.157         & 61.759          & 92.256          & 55.759          \\
		ISTDU-Net \citep{ref31}                  & 75.674          & 95.038          & 30.333          & 91.321          & 98.307          & 7.514          & 63.334          & 93.266          & 29.986          \\
		SCTransNet \citep{ref6}                & 75.024          & {\ul 95.420}    & 39.157          & {\ul 92.992}    & {\ul 98.730}    & 3.102          & 63.997          & {\ul 94.276}    & {\ul 18.314}    \\
		LCAE-Net                       & \textbf{80.421} & \textbf{96.565} & \textbf{11.720} & \textbf{94.746} & \textbf{99.259} & \textbf{1.034} & \textbf{70.730} & \textbf{95.286} & 19.017          \\ \hline
	\end{tabular}
	\vspace{-\baselineskip}
\end{table}
\par We can see from the \cref{table1} that LCAE-Net outperforms the other SOTA methods and has achieved excellent test results on three benchmark datasets. Specifically, in terms of $IoU$ metric, LCAE-Net has reached 80.421\%, 94.746\% and 70.730\% respectively on the three datasets, outperforming the second-ranked method by 3.651\%, 1.742\% and 6.585\%, demonstrating a significant advantage over the second-ranked method. In terms of ${P_d}$, LCAE-Net has achieved 96.565\%, 99.259\% and 95.286\% on three datasets, surpassing the second-ranked method by 0.764\%, 1.742\%, and 1.01\% respectively. As for ${F_a}$, LCAE-Net excels on the NUAA-SIRST and NUDT-SIRST datasets with 11.720$\times{10^{ - 6}}$ and 1.034$\times{10^{ - 6}}$ separately, but is marginally less effective than DNA-Net and SCTransNet on the IRSTD-1K dataset. The dense nested interactive connection of DNA-Net and the Spatial-channel Cross Transformer Block (SCTB) of SCTransNet facilitate multiple enhancements of semantic information at the expense of computational efficiency, which make them have certain advantages when confronted with datasets like IRSTD-1K, which including varying-shape targets and background with clutters and noises. Overall, our method exhibits favorable results across the three datasets. Additionally, \cref{table1} indicates that DNA-Net, UIU-Net, and SCTransNet exhibit strong competitiveness with our method in these three metrics. \cref{table2} presents a comparison of these methods in terms of model parameter size, FLOPs and detection speed, calculated using images with a resolution of 256×256, where the optimal value in each column is highlighted in \textbf{bold}.
\begin{table}[H]
	\centering
	\caption{Comparison of model complexity and detection speed, where the optimal value in each column is highlighted in \textbf{bold}.}
	\label{table2}
	\begin{tabular}{cccc}
		\hline
		Method     & Params (M) & FLOPs (G) & Speed (fps)   \\ \hline
		DNA-Net \citep{ref30}   & 4.697      & 14.261    & 40.49 \\
		UIU-Net  \citep{ref33}  & 50.540     & 54.426    & 54.17 \\
		SCTransNet  \citep{ref6}& 11.191     & 10.119    & 33.62 \\
		LCAE-Net   & \textbf{1.945}      & \textbf{4.862}     & \textbf{70.23} \\ \hline
	\end{tabular}
	\vspace{-\baselineskip}
\end{table}
\par
\begin{figure*}[htp]\centering
	\includegraphics[width=0.84\linewidth]{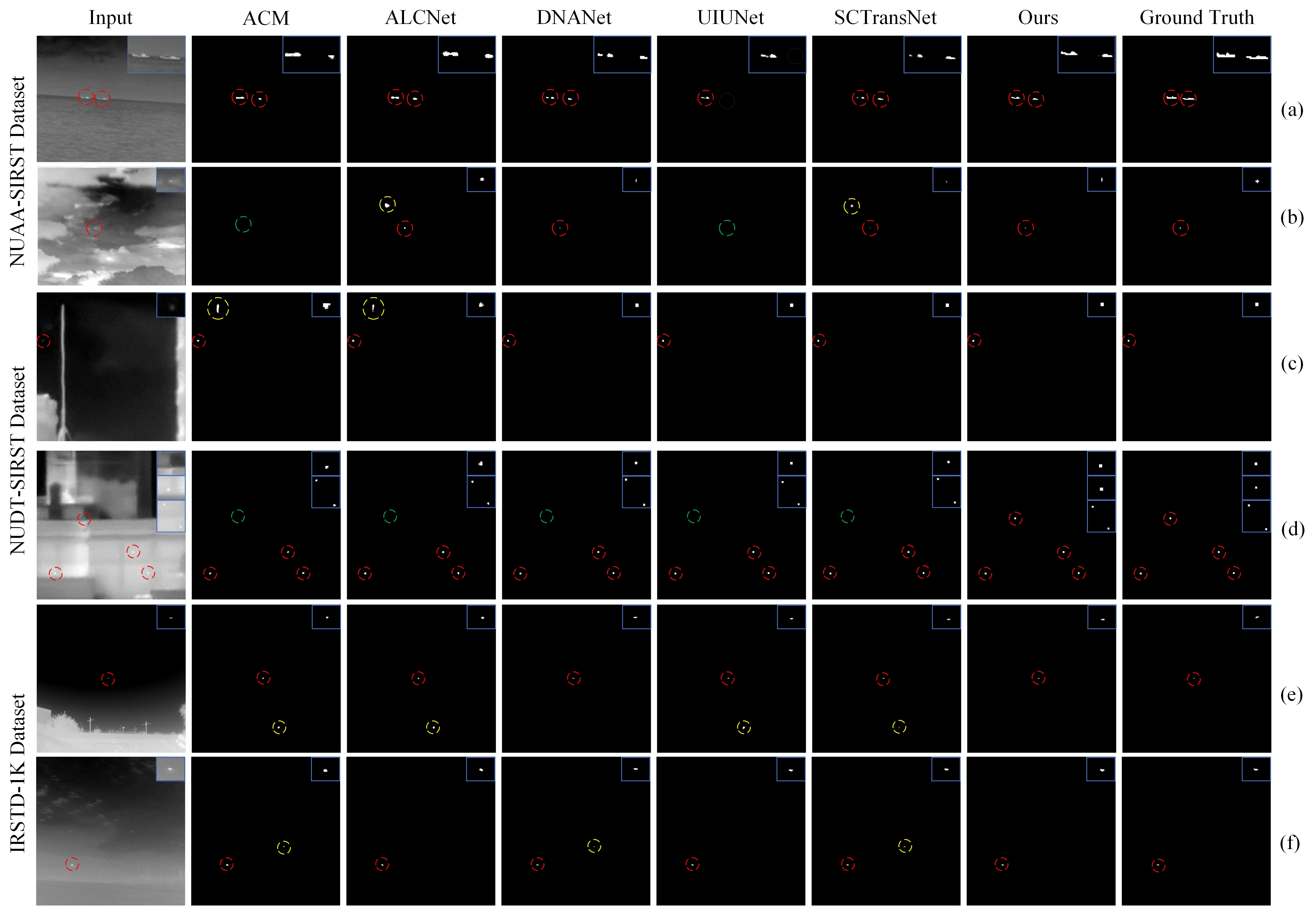}
	\caption{Visual results obtained by different IRSTD methods. The red, green, and yellow circles represent correctly detected targets, missed detection, and false alarms respectively.}
	\vspace{-\baselineskip}
	\label{fig_6}
\end{figure*}

\begin{figure*}[htp]
	\centering
	\includegraphics[width=0.84\linewidth]{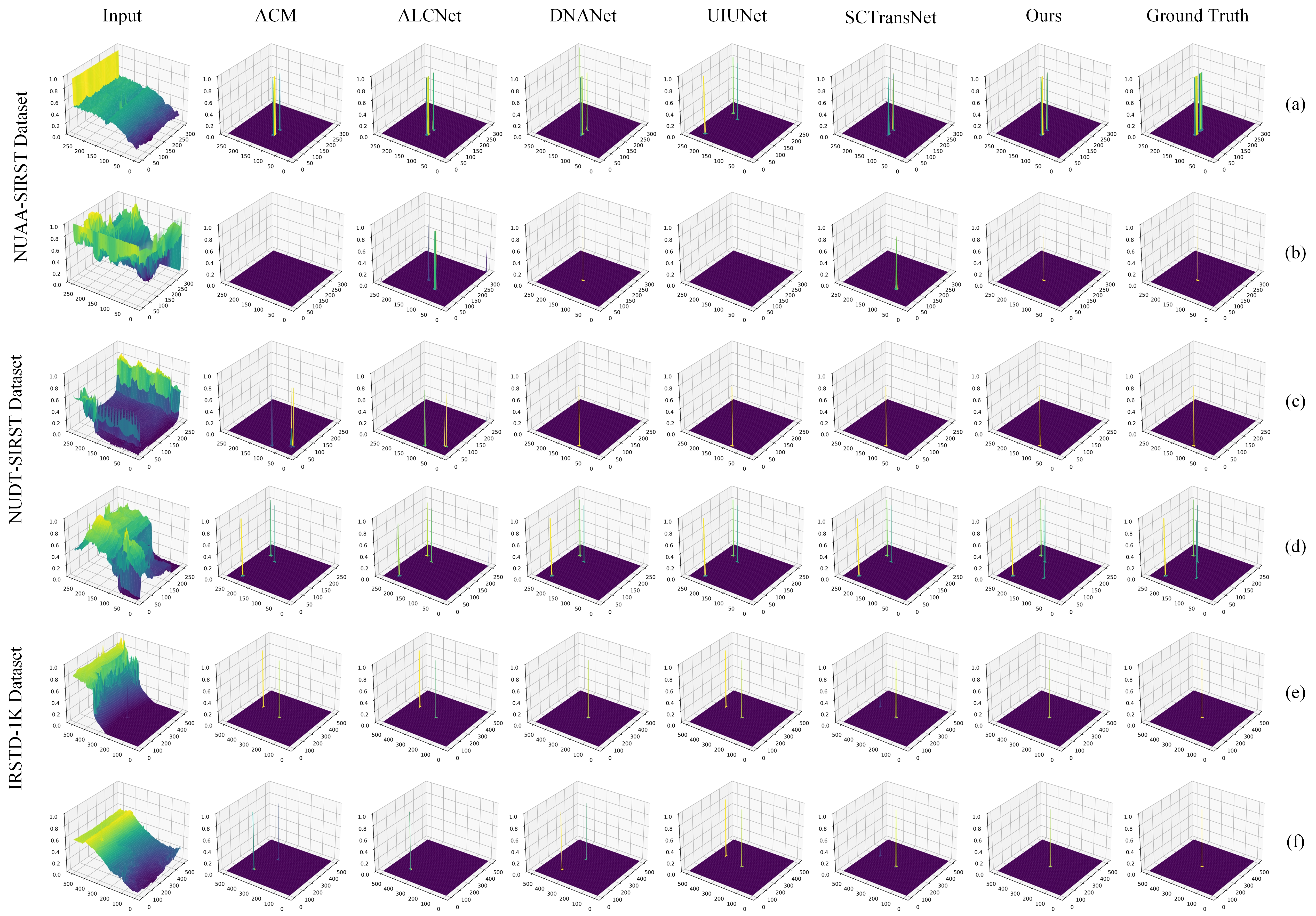}
	\caption{3D visualization of salient maps of different methods on six test images.}
	\label{fig_7}
	\vspace{-\baselineskip}
\end{figure*}

 Considering the complexity and performance of each method, although our LCAE-Net has a small gap in ${F_a}$ on the IRSTD-1K dataset, its parameters, FLOPs and detection speed are 1.945M, 4.862G and 70.23fps respectively, representing a substantial computation cost reduction and high real-time performance compared to competitive methods. This indicates that our LCAE-Net effectively balances the demands for computational complexity and detection performance, making it more suitable for deployment on edge infrared detection devices with limited resources. To visually evaluate the differences in detection effects among these methods, we selects two typical infrared scenarios from each of the three public datasets and compares the detection results of six detection methods. The visualizations and corresponding saliency maps are shown in \cref{fig_6} and \cref{fig_7}, while in \cref{fig_6}, the red, green, and yellow circles represent correctly detected targets, missed detection, and false alarms respectively.
\par From two figures we can observe that our method has achieved satisfactory detection results across all six scenarios. In scenario (a), the original image depicts two closely adjacent ships on the sea. Our LCAE-Net accurately detects the shape of two ships. In contrast, except for ACM and ALCNet, the other methods detect the left target as two separate entities, demonstrating that the our method is more suitable for detecting strip-like targets. It is worth noting that comparing to ALCNet which uses similar multiscale local contrast enhancing method, our detection result is more precise, which indicates a single-scale operator with appropriate prior knowledge is sufficient to produce satisfactory results. Scenarios (b) and (f) are infrared small targets in the sky. In scenario (b), ACM and UIUNet miss targets, while ALCNet and SCTransNet have false alarms. In scenario (f), ACM, DNANet, and SCTransNet all have false alarms. Our LCAE-Net achieves better detection results in both scenarios, indicating that the proposed LCE module could guide the network model to pay more attention on real positions of small targets. In scenario (c), the tail flame of the rocket is misdetected by ACM and ALCNet, and in scenario (e), the street lamp is misdetected by ACM, ALCNet, UIUNet, and SCTransNet. Our method exhibits good detection performance in these two scenarios, highlighting its effectiveness in suppressing irrelevant background interference. In scenario (d), all other methods except the one we proposed miss targets, indicating that even dim targets can be effectively detected with the assistance of appropriate prior knowledge.
\par
\begin{figure}
	\centering
	\includegraphics[width=\linewidth]{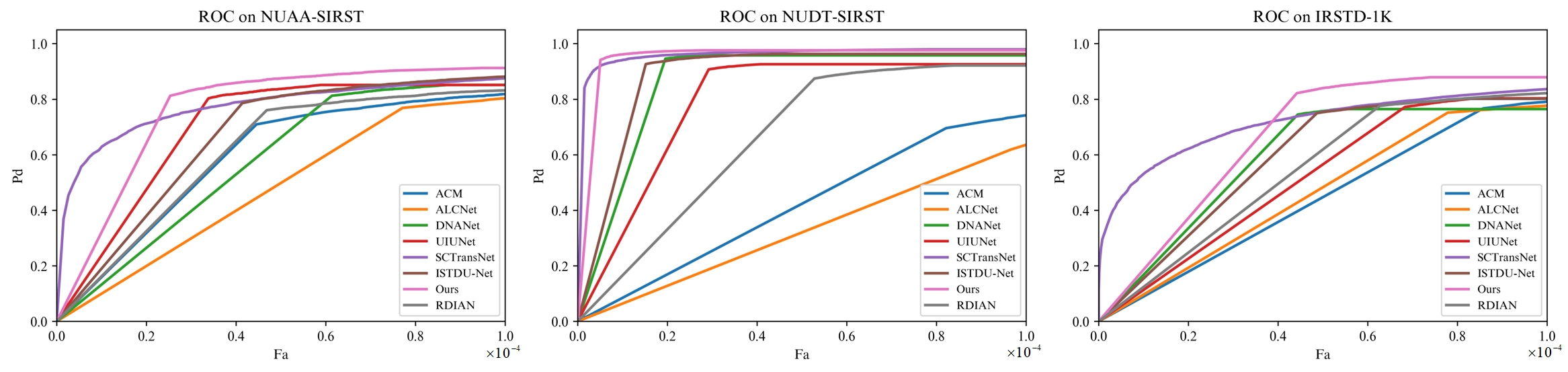}
	\caption{ROC curves of different methods on three datasets.}
	\label{fig_8}
	\vspace{-\baselineskip}
\end{figure}

 \cref{fig_8} shows the Receiver Operating Charateristic (ROC) curves of different methods across three datasets. A ROC curve reflects the detection capability of one method. From \cref{fig_8}, it could be seen that our LCAE-Net achieve a high detection probability at a very low false alarm rate, proving the superiority of it.
\subsection{Ablation study}
\par In order to demonstrate the effectiveness of our major modules, we conducted ablation experiments on three benchmark datasets using the same implementation details in Section \ref{Implementation details}. The experimental results are shown in \cref{table3}, and the optimal value in each column is highlighted in \textbf{bold}.
\begin{table}[htp]
	\centering
	\caption{Results of ablation experiments. The optimal value in each column is highlighted in \textbf{bold}.}
	\label{table3}
	\begin{tabular}{ccccccccccc}
		\hline
		\multicolumn{1}{l}{\multirow{2}{*}{CAE}} & \multicolumn{1}{l}{\multirow{2}{*}{LCE}} & \multicolumn{3}{c}{NUAA-SIRST}                      & \multicolumn{3}{c}{NUDT-SIRST}                     & \multicolumn{3}{c}{IRSTD-1K}                        \\ \cline{3-11} 
		\multicolumn{1}{l}{}                             & \multicolumn{1}{l}{}                      & $IoU$/\%         & ${P_d}$/\%           & ${F_a}$/${10^{ - 6}}$         & $IoU$/\%         & ${P_d}$/\%           & ${F_a}$/${10^{ - 6}}$        & $IoU$/\%         & ${P_d}$/\%           & ${F_a}$/${10^{ - 6}}$         \\ \hline
		&                                           & 77.733          & 94.656          & 22.336          & 92.887          & 98.836          & 2.918          & 67.250          & 94.613          & 54.620          \\
		$\checkmark$                                                &                                           & 78.465          & 96.183          & 29.782          & 93.122          & 99.048          & 2.827          & 68.679          & 94.613          & 30.986          \\
		& $\checkmark$                                         & 78.847          & \textbf{96.565}          & 21.233          & 93.729          & 99.048          & 2.735          & 68.342          & 94.949          & 35.243          \\
		$\checkmark$                                                & $\checkmark$                                         & \textbf{80.421} & \textbf{96.565} & \textbf{11.720} & \textbf{94.746} & \textbf{99.259} & \textbf{1.034} & \textbf{70.730} & \textbf{95.286} & \textbf{19.017} \\ \hline
	\end{tabular}
	\vspace{-\baselineskip}
\end{table}
\par The first line of results in \cref{table3} represent the network after replacing CAE module with element-wise summation and removing LCE module. Comparing the first and second lines, it becomes apparent that integrating the proposed CAE module enhances the network's performance, resulting in improvements of 0.732\%, 0.235\%, and 1.429\% in $IoU$ across the three datasets. Additionally, ${P_d}$ metric increases by 1.527\% and 0.212\% on the NUAA-SIRST and NUDT-SIRST datasets separately, while ${F_a}$ metric decreases by 0.091×${10^{ - 6}}$ and 23.634×${10^{ - 6}}$ on the NUDT-SIRST and IRSTD-1K datasets, validating the efficacy of the proposed CAE module. Furthermore, a comparison between the first and third line highlights improvements in all evaluation metrics on the three datasets, with the ${P_d}$ metric on the NUAA-SIRST dataset achieving the optimal value in its column, thereby confirming the effectiveness of the developed LCE module. Lastly, comparing to integrate only one module, data from the first to the fourth lines presents the remarkable impact of utilizing the CAE and LCE modules simultaneously. This combination leads to optimal results across all metrics, emphasizing the substantial boost in network performance delivered by our proposed modules.
\subsection{Hyperparameter Analysis}\label{Hyperparameter Analysis}
\begin{table*}
	\centering
	\caption{Analysis of the hyperparameters. The optimal value and the suboptimal value in each column are highlighted in \textbf{bold} and \underline{underline} respectively.}
	\label{table4}
	\begin{tabular}{cccccccccccc}
		\hline
		\multirow{2}{*}{$d$} & \multirow{2}{*}{$\alpha$} & \multirow{2}{*}{$\beta$} & \multicolumn{3}{c}{NUAA-SIRST}                      & \multicolumn{3}{c}{NUDT-SIRST}                     & \multicolumn{3}{c}{IRSTD-1K}                        \\ \cline{4-12} 
		&                    &                    & $IoU$/\%         & ${P_d}$/\%           & ${F_a}$/${10^{ - 6}}$         & $IoU$/\%         & ${P_d}$/\%           & ${F_a}$/${10^{ - 6}}$        & $IoU$/\%         & ${P_d}$/\%           & ${F_a}$/${10^{ - 6}}$         \\ \hline
		\multirow{6}{*}{1} & 1                  & 0.5                & \textbf{80.421} & \textbf{96.565} & \textbf{11.720} & \textbf{94.746} & \textbf{99.259} & \textbf{1.034} & \textbf{70.730} & \textbf{95.286} & \textbf{19.017} \\
		& 1                  & 1                  & 78.447          & 95.802          & 26.059          & {\ul 93.863}    & {\ul 99.153}    & 2.964          & 66.818          & 94.613          & 32.207          \\
		& 1.5                & 0.5                & 78.248          & {\ul 96.183}    & 25.507          & 93.165          & 98.519          & 1.999          & 67.787          & {\ul 94.949}    & 42.322          \\
		& 1.5                & 1                  & 77.263          & 95.038          & 24.335          & 93.172          & {\ul 99.153}    & 4.803          & 67.195          & 92.929          & 41.848          \\
		& 2                  & 0.5                & 78.113          & 93.130          & 28.748          & 93.525          & 98.836          & 3.309          & 66.036          & 94.276          & 48.737          \\
		& 2                  & 1                  & 77.044          & \textbf{96.565} & 23.232          & 93.078          & 98.942          & 3.815          & 65.950          & 93.939          & 51.166          \\ \hline
		\multirow{6}{*}{2} & 1                  & 0.5                & {\ul 79.551}    & \textbf{96.565} & {\ul 12.409}    & 93.019          & 98.730          & 1.655          & 68.538          & 93.939          & 38.887          \\
		& 1                  & 1                  & 77.541          & 95.038          & 23.784          & 93.221          & \textbf{99.259} & 3.746          & 66.196          & {\ul 94.949}    & 42.170          \\
		& 1.5                & 0.5                & 77.510          & 94.275          & 36.882          & 93.195          & 98.624          & 1.861          & 66.083          & 93.266          & {\ul 29.701}    \\
		& 1.5                & 1                  & 77.112          & 95.420          & 38.123          & 92.339          & 98.307          & 4.665          & 67.697          & 94.613          & 34.370          \\
		& 2                  & 0.5                & 79.535          & {\ul 96.183}    & 19.785          & 93.722          & 98.942          & 2.895          & 64.316          & 93.603          & 41.639          \\
		& 2                  & 1                  & 77.735          & 95.802          & 35.779          & 92.688          & 98.834          & 3.907          & 66.144          & 94.276          & 43.632          \\ \hline
		\multirow{6}{*}{3} & 1                  & 0.5                & 79.347          & \textbf{96.565} & 29.782          & 93.172          & 98.941          & 2.528          & {\ul 68.706}    & 93.939          & 36.439          \\
		& 1                  & 1                  & 77.059          & 93.893          & 37.434          & 92.941          & 98.730          & 2.045          & 68.395          & 94.276          & 31.239          \\
		& 1.5                & 0.5                & 78.492          & 94.656          & 28.748          & 93.016          & 98.519          & 2.436          & 68.581          & {\ul 94.949}    & 36.610          \\
		& 1.5                & 1                  & 78.410          & 95.802          & 32.263          & 93.841          & 98.836          & 1.976          & 68.123          & 94.613          & 35.907          \\
		& 2                  & 0.5                & 79.025          & {\ul 96.183}    & 41.984          & 92.909          & 98.413          & 3.700          & 67.222          & 93.602          & 34.731          \\
		& 2                  & 1                  & 77.679          & 95.038          & 34.814          & 93.172          & 98.624          & 2.666          & 66.253          & \textbf{95.286} & 45.890          \\ \hline
		\multirow{6}{*}{4} & 1                  & 0.5                & 79.053          & {\ul 96.183}    & 40.743          & 92.733          & 98.413          & {\ul 1.310}    & 67.288          & \textbf{95.286} & 46.421          \\
		& 1                  & 1                  & 78.903          & 95.802          & 30.678          & 92.291          & 98.836          & 4.895          & 68.849          & {\ul 94.949}    & 34.086          \\
		& 1.5                & 0.5                & 76.444          & 95.038          & 19.579          & 93.730          & {\ul 99.153}    & 1.563          & 67.118          & 95.623          & 32.017          \\
		& 1.5                & 1                  & 78.023          & \textbf{96.565} & 22.060          & 92.860          & 98.942          & 5.079          & 68.234          & 94.276          & 39.779          \\
		& 2                  & 0.5                & 78.121          & 94.656          & 29.161          & 93.016          & 98.624          & 1.792          & 67.946          & 94.613          & 35.870          \\
		& 2                  & 1                  & 78.872          & 95.420          & 28.196          & 92.859          & 98.519          & 3.723          & 68.121          & 94.613          & 43.423          \\ \hline
	\end{tabular}
	\vspace{-\baselineskip}
\end{table*}
\par The LCE module incorporates two hyperparameters $\alpha$ and $\beta$, with the objective of emphasizing the pixels at the potential infrared small target in the image. Inspired by the design of fixed convolution operators utilized in methods such as RDIAN \citep{ref5} and MPCM \citep{ref24}, it is reasonable to assume that the values of $\alpha$ and $\beta$ should not exhibit significant disparities. Therefore, the value range for hyperparameter $\alpha$ is designated at $\alpha  \in \{ 1,1.5,2\} $, while for $\beta$, the value range is set as $\beta  \in \{ 0.5,1\} $. Given that infrared small targets generally do not exceed in size $9 \times 9$ within the image, and considering that the convolution kernel size should be an odd number, the value range for $d$ is set as $d \in \{ 1,2,3,4\} $. The experimental results under different hyperparameter settings are displayed in \cref{table4}, where the optimal value and the suboptimal value in each column are highlighted in \textbf{bold} and \underline{underline} respectively.

\begin{table}
	\centering
	\caption{Analysis of the refined hyperparameters. The optimal value and the worst value in each column are highlighted in \textbf{bold} and \underline{underline} respectively.}
	\label{table5}
	\begin{tabular}{ccccccccccc}
		\hline
		\multirow{2}{*}{$\alpha$} & \multirow{2}{*}{$\beta$} & \multicolumn{3}{c}{NUAA-SIRST}                      & \multicolumn{3}{c}{NUDT-SIRST}                     & \multicolumn{3}{c}{IRSTD-1K}                        \\ \cline{3-11} 
		&                    & $IoU$/\%         & ${P_d}$/\%           & ${F_a}$/${10^{ - 6}}$         & $IoU$/\%         & ${P_d}$/\%           & ${F_a}$/${10^{ - 6}}$        & $IoU$/\%         & ${P_d}$/\%           & ${F_a}$/${10^{ - 6}}$         \\ \hline
		0.5                & 0.25               & 78.402          & \textbf{96.565}          & 22.060          & 93.791          & 99.153    & 2.367          & 68.595    & 94.613          & 31.542    \\
		0.5                & 0.5                & 78.057          & 95.420          & 18.269          & 93.698          & 98.942          & 2.068          & 67.522          & 94.949   & 33.023          \\
		0.5                & 0.75               & {\ul 76.131}          & 94.656          & 32.884         & 93.185          & {\ul 98.413}          & 4.964          & 67.211          & 94.613          & 38.697          \\
		0.5                & 1                  & 77.867          & 94.274          & 30.057          & 93.095          & 98.730          & 2.620          & 65.983          & 93.266          & 34.199          \\ \hline
		1                  & 0.25               & 77.774          & 93.893          & 28.748          & 93.023          & 98.836          & 3.654          & 67.760          & 93.603          & 30.252          \\
		1                  & 0.5                & \textbf{80.421} & \textbf{96.565} & \textbf{11.720} & \textbf{94.746} & \textbf{99.259} & \textbf{1.034} & \textbf{70.730} & \textbf{95.286} & \textbf{19.017} \\
		1                  & 0.75               & 78.492          & 95.420          & 28.127          & 93.571          & 98.730          & 1.792          & 65.445          & \textbf{95.286} & 40.178          \\
		1                  & 1                  & 78.447          & 95.802          & 26.059          & 93.321          & 98.836          & 3.516          & 66.818          & 94.613          & 32.207          \\ \hline
		1.5                & 0.25               & 78.089          & 94.656          & 31.574          & 93.277          & 98.624          & 1.746          & 66.507          & 94.613          & 32.700          \\
		1.5                & 0.5                & 78.248          & 96.183    & 25.507          & 93.165          & 98.519          & 1.999          & 67.787          & 94.949    & 42.322          \\
		1.5                & 0.75               & 77.702          & 95.802          & 25.163          & 93.152          & 99.153          & {\ul 5.377}          & 67.311          & 94.613          & 33.592          \\
		1.5                & 1                  & 77.263          & 95.038          & 24.335          & 93.172          & 99.153          & 4.803          & 67.195          & {\ul 92.929}          & 41.848          \\ \hline
		2                  & 0.25               & 76.824          & 94.275          & {\ul 33.849}          & {\ul 92.949}          & 98.730          & 3.102          & 66.975          & 93.603    & 37.046          \\
		2                  & 0.5                & 78.113          & {\ul 93.130}          & 28.748          & 93.525          & 98.836          & 3.309          & 66.036          & 94.276          & 48.737          \\
		2                  & 0.75               & 78.540          & \textbf{96.565} & 21.647    & 93.824          & 98.519          & 2.964          & {\ul 64.161}          & 94.613          & 37.331          \\
		2                  & 1                  & 77.044          & \textbf{96.565} & 23.232          & 93.078          & 98.942          & 3.815          & 65.950          & 93.939          & {\ul 51.166}          \\ \hline
		\multicolumn{2}{c}{\textbf{$distance$}} &4.290 &3.435 &22.129 &1.797 &0.846 &4.343 &6.569 &2.357 &32.149 \\ \hline
	\end{tabular}
\end{table}
\par We can see from the \cref{table4} that when $d$ is fixed, LCAE-Net performs well on the three datasets when $\alpha$ and $\beta$ are set to 1 and 0.5, and the global optimal value is acquired when $d$ is set to 1. We conclude that for $d$, due to the generally small size of infrared small targets, when the value is too large, the convolution operator covers an excessive amount of non-target area. This leads to interactions between the center pixel and distant, irrelevant pixels, ultimately making it challenging to enhance performance. To further explore the impact of $\alpha$ and $\beta$, we fixed $d$ at 1 and refined the value range of $\alpha$ and $\beta$. Here the value range for hyperparameter $\alpha$ is designated at $\alpha  \in \{ 0.5,1,1.5,2\} $, while for $\beta$, the value range is set as $\beta  \in \{ 0.25,0.5,0.75,1\} $. The experimental results on the three datasets are shown in \cref{table5}. The optimal value and the worst value in each column are highlighted in \textbf{bold} and \underline{underline} respectively, and the last line shows the distance between the two vaules.
\par As evident from the data presented in the \cref{table5}, LCAE-Net exhibits the strongest robustness on ${P_d}$ metric, with distances across the three datasets being 3.435\%, 0.846\%, and 2.357\% respectively, indicating its effectiveness in detecting infrared small targets in images. In terms of $IoU$ and ${F_a}$ metrics, the method performs well on the NUAA-SIRST and NUDT-SIRST datasets, with extreme differences of 4.290\%, 22.129$\times{10^{ - 6}}$, and 1.797\%, 4.343$\times{10^{ - 6}}$ respectively. However, on the IRSTD-1K dataset, the robustness is slightly weaker, with extreme differences of 6.569\% and 32.149$\times{10^{ - 6}}$ for the two metrics. This is because the IRSTD-1K dataset contains a diverse range of infrared small targets across various scenes such as oceans, rivers, fields, mountains, cities, and clouds. The relatively severe clutter and noise in these scenes render this dataset highly sensitive to hyperparameter values. This could explain the slightly weaker performance of the proposed method on the IRSTD-1K dataset. Additionally, we can see from the table that when the difference between values of $\alpha$ and $\beta$ is too large (see $\alpha$=2 and $\beta$=0.25), LCAE-Net may perform weakly. This may due to the fact that when the difference is too large, the calculated $LCD$ value consistently exhibits a large negative absolute value, thus lead to the large $LCA$ value and decrease the discrimination. In this situation, whether the convolution operator passes through a target pixel or a background pixel, it will smooth the entire image and reducing the disparity in grayscale values between pixels on the original image, making it difficult for the detection network to effectively identify the target. When we choose hyperparameters in practice, it is advisable for us not allow for a large gap between these two values.
\section{Conclusion}
\par This paper proposes an infrared small target detection method called LCAE-Net that seamlessly integrates model-driven and data-driven approaches, incorporating prior knowledge into the neural network's learning and distinguishing process. In this model, we design the LCE module to guide the neural network's focus towards the spatial location of infrared small targets, and propose the CAE module to efficiently fuse and enhance information across different channels. Experimental results indicate that the proposed method has achieved favorable outcomes on three publicly available infrared small target datasets. Furthermore, its parameter size and FLOPs are relatively small, making it suitable for deployment on edge devices with limited computational resources.
\par The LCAE-Net proposed in our paper has achieved a commendable balance between detection performance and computational cost. There still exists improving space in false alarm rate. In future, we will continuously optimizing the model to decrease its false alarm rate, and practical deploy it on low-resource devices.
	\appendix
	
	\printcredits
	
	\bibliographystyle{cas-model2-names}
	
	\bibliography{cas-refs}

\begin{thebibliography}{56}
\expandafter\ifx\csname natexlab\endcsname\relax\def\natexlab#1{#1}\fi
\providecommand{\url}[1]{\texttt{#1}}
\providecommand{\href}[2]{#2}
\providecommand{\path}[1]{#1}
\providecommand{\DOIprefix}{doi:}
\providecommand{\ArXivprefix}{arXiv:}
\providecommand{\URLprefix}{URL: }
\providecommand{\Pubmedprefix}{pmid:}
\providecommand{\doi}[1]{\href{http://dx.doi.org/#1}{\path{#1}}}
\providecommand{\Pubmed}[1]{\href{pmid:#1}{\path{#1}}}
\providecommand{\bibinfo}[2]{#2}
\ifx\xfnm\relax \def\xfnm[#1]{\unskip,\space#1}\fi
%Type = Article
\bibitem[{Bi et~al.(2019)Bi, Chen, Sun and Bai}]{ref11}
\bibinfo{author}{Bi, Y.}, \bibinfo{author}{Chen, J.}, \bibinfo{author}{Sun,
  H.}, \bibinfo{author}{Bai, X.}, \bibinfo{year}{2019}.
\newblock \bibinfo{title}{Fast detection of distant, infrared targets in a
  single image using multiorder directional derivatives}.
\newblock \bibinfo{journal}{IEEE Transactions on Aerospace and Electronic
  Systems} \bibinfo{volume}{56}, \bibinfo{pages}{2422--2436}.
\newblock \DOIprefix\doi{10.1109/TAES.2019.2946678}.
%Type = Inproceedings
\bibitem[{Chapple et~al.(1999)Chapple, Bertilone, Caprari, Angeli and
  Newsam}]{ref2}
\bibinfo{author}{Chapple, P.B.}, \bibinfo{author}{Bertilone, D.C.},
  \bibinfo{author}{Caprari, R.S.}, \bibinfo{author}{Angeli, S.},
  \bibinfo{author}{Newsam, G.N.}, \bibinfo{year}{1999}.
\newblock \bibinfo{title}{Target detection in infrared and sar terrain images
  using a non-gaussian stochastic model}, in: \bibinfo{booktitle}{Targets and
  backgrounds: characterization and representation V},
  \bibinfo{organization}{SPIE}. pp. \bibinfo{pages}{122--132}.
%Type = Article
\bibitem[{Chen et~al.(2013)Chen, Li, Wei, Xia and Tang}]{ref22}
\bibinfo{author}{Chen, C.P.}, \bibinfo{author}{Li, H.}, \bibinfo{author}{Wei,
  Y.}, \bibinfo{author}{Xia, T.}, \bibinfo{author}{Tang, Y.Y.},
  \bibinfo{year}{2013}.
\newblock \bibinfo{title}{A local contrast method for small infrared target
  detection}.
\newblock \bibinfo{journal}{IEEE transactions on geoscience and remote sensing}
  \bibinfo{volume}{52}, \bibinfo{pages}{574--581}.
\newblock \DOIprefix\doi{10.1109/TGRS.2013.2242477}.
%Type = Article
\bibitem[{Chen et~al.(2022)Chen, Li, Liu and Su}]{ref47}
\bibinfo{author}{Chen, Y.}, \bibinfo{author}{Li, L.}, \bibinfo{author}{Liu,
  X.}, \bibinfo{author}{Su, X.}, \bibinfo{year}{2022}.
\newblock \bibinfo{title}{A multi-task framework for infrared small target
  detection and segmentation}.
\newblock \bibinfo{journal}{IEEE Transactions on Geoscience and Remote Sensing}
  \bibinfo{volume}{60}, \bibinfo{pages}{1--9}.
\newblock \DOIprefix\doi{10.1109/TGRS.2022.3195740}.
%Type = Article
\bibitem[{Cui et~al.(2024)Cui, Lei, Chen, Zhang, Peng, Hao and Zhang}]{ref14}
\bibinfo{author}{Cui, Y.}, \bibinfo{author}{Lei, T.}, \bibinfo{author}{Chen,
  G.}, \bibinfo{author}{Zhang, Y.}, \bibinfo{author}{Peng, L.},
  \bibinfo{author}{Hao, X.}, \bibinfo{author}{Zhang, G.}, \bibinfo{year}{2024}.
\newblock \bibinfo{title}{Hollow side window filter with saliency prior for
  infrared small target detection}.
\newblock \bibinfo{journal}{IEEE Geoscience and Remote Sensing Letters}
  \bibinfo{volume}{21}, \bibinfo{pages}{1--5}.
\newblock \DOIprefix\doi{10.1109/LGRS.2023.3342981}.
%Type = Article
\bibitem[{Dai and Wu(2017)}]{ref18}
\bibinfo{author}{Dai, Y.}, \bibinfo{author}{Wu, Y.}, \bibinfo{year}{2017}.
\newblock \bibinfo{title}{Reweighted infrared patch-tensor model with both
  nonlocal and local priors for single-frame small target detection}.
\newblock \bibinfo{journal}{IEEE journal of selected topics in applied earth
  observations and remote sensing} \bibinfo{volume}{10},
  \bibinfo{pages}{3752--3767}.
\newblock \DOIprefix\doi{10.1109/JSTARS.2017.2700023}.
%Type = Article
\bibitem[{Dai et~al.(2016)Dai, Wu and Song}]{ref16}
\bibinfo{author}{Dai, Y.}, \bibinfo{author}{Wu, Y.}, \bibinfo{author}{Song,
  Y.}, \bibinfo{year}{2016}.
\newblock \bibinfo{title}{Infrared small target and background separation via
  column-wise weighted robust principal component analysis}.
\newblock \bibinfo{journal}{Infrared Physics \& Technology}
  \bibinfo{volume}{77}, \bibinfo{pages}{421--430}.
\newblock \DOIprefix\doi{10.1016/j.infrared.2016.06.021}.
%Type = Inproceedings
\bibitem[{Dai et~al.(2021a)Dai, Wu, Zhou and Barnard}]{ref29}
\bibinfo{author}{Dai, Y.}, \bibinfo{author}{Wu, Y.}, \bibinfo{author}{Zhou,
  F.}, \bibinfo{author}{Barnard, K.}, \bibinfo{year}{2021}a.
\newblock \bibinfo{title}{Asymmetric contextual modulation for infrared small
  target detection}, in: \bibinfo{booktitle}{Proceedings of the IEEE/CVF winter
  conference on applications of computer vision}, pp.
  \bibinfo{pages}{950--959}.
\newblock \DOIprefix\doi{10.1109/WACV48630.2021.00099}.
%Type = Article
\bibitem[{Dai et~al.(2021b)Dai, Wu, Zhou and Barnard}]{ref7}
\bibinfo{author}{Dai, Y.}, \bibinfo{author}{Wu, Y.}, \bibinfo{author}{Zhou,
  F.}, \bibinfo{author}{Barnard, K.}, \bibinfo{year}{2021}b.
\newblock \bibinfo{title}{Attentional local contrast networks for infrared
  small target detection}.
\newblock \bibinfo{journal}{IEEE transactions on geoscience and remote sensing}
  \bibinfo{volume}{59}, \bibinfo{pages}{9813--9824}.
\newblock \DOIprefix\doi{10.1109/TGRS.2020.3044958}.
%Type = Inproceedings
\bibitem[{Deshpande et~al.(1999)Deshpande, Er, Venkateswarlu and Chan}]{ref10}
\bibinfo{author}{Deshpande, S.D.}, \bibinfo{author}{Er, M.H.},
  \bibinfo{author}{Venkateswarlu, R.}, \bibinfo{author}{Chan, P.},
  \bibinfo{year}{1999}.
\newblock \bibinfo{title}{Max-mean and max-median filters for detection of
  small targets}, in: \bibinfo{booktitle}{Signal and Data Processing of Small
  Targets 1999}, \bibinfo{organization}{SPIE}. pp. \bibinfo{pages}{74--83}.
%Type = Article
\bibitem[{Gao et~al.(2013)Gao, Meng, Yang, Wang, Zhou and Hauptmann}]{ref15}
\bibinfo{author}{Gao, C.}, \bibinfo{author}{Meng, D.}, \bibinfo{author}{Yang,
  Y.}, \bibinfo{author}{Wang, Y.}, \bibinfo{author}{Zhou, X.},
  \bibinfo{author}{Hauptmann, A.G.}, \bibinfo{year}{2013}.
\newblock \bibinfo{title}{Infrared patch-image model for small target detection
  in a single image}.
\newblock \bibinfo{journal}{IEEE transactions on image processing}
  \bibinfo{volume}{22}, \bibinfo{pages}{4996--5009}.
\newblock \DOIprefix\doi{10.1109/TIP.2013.2281420}.
%Type = Inproceedings
\bibitem[{Girshick et~al.(2014)Girshick, Donahue, Darrell and Malik}]{ref53}
\bibinfo{author}{Girshick, R.}, \bibinfo{author}{Donahue, J.},
  \bibinfo{author}{Darrell, T.}, \bibinfo{author}{Malik, J.},
  \bibinfo{year}{2014}.
\newblock \bibinfo{title}{Rich feature hierarchies for accurate object
  detection and semantic segmentation}, in: \bibinfo{booktitle}{Proceedings of
  the IEEE/CVF conference on computer vision and pattern recognition}, pp.
  \bibinfo{pages}{580--587}.
%Type = Article
\bibitem[{He et~al.(2024)He, Pan and An}]{ref26}
\bibinfo{author}{He, S.}, \bibinfo{author}{Pan, S.}, \bibinfo{author}{An, B.},
  \bibinfo{year}{2024}.
\newblock \bibinfo{title}{Infrared small target detection based on variance
  difference weighted three-layer local contrast measure}.
\newblock \bibinfo{journal}{Infrared Physics \& Technology}
  \bibinfo{volume}{139}, \bibinfo{pages}{105315}.
\newblock \DOIprefix\doi{10.1016/j.infrared.2024.105315}.
%Type = Article
\bibitem[{Hou et~al.(2022)Hou, Zhang, Tan, Xi, Zheng and Li}]{ref31}
\bibinfo{author}{Hou, Q.}, \bibinfo{author}{Zhang, L.}, \bibinfo{author}{Tan,
  F.}, \bibinfo{author}{Xi, Y.}, \bibinfo{author}{Zheng, H.},
  \bibinfo{author}{Li, N.}, \bibinfo{year}{2022}.
\newblock \bibinfo{title}{Istdu-net: Infrared small-target detection u-net}.
\newblock \bibinfo{journal}{IEEE Geoscience and Remote Sensing Letters}
  \bibinfo{volume}{19}, \bibinfo{pages}{1--5}.
\newblock \DOIprefix\doi{10.1109/LGRS.2022.3141584}.
%Type = Article
\bibitem[{Hu et~al.(2020)Hu, Shen and Sun}]{ref41}
\bibinfo{author}{Hu, J.}, \bibinfo{author}{Shen, L.}, \bibinfo{author}{Sun,
  G.}, \bibinfo{year}{2020}.
\newblock \bibinfo{title}{Squeeze-and-excitation networks}.
\newblock \bibinfo{journal}{IEEE Transactions on Pattern Analysis and Machine
  Intelligence} \bibinfo{volume}{42}, \bibinfo{pages}{2011--2023}.
\newblock \DOIprefix\doi{10.1109/TPAMI.2019.2913372}.
%Type = Article
\bibitem[{Hu et~al.(2024)Hu, Yang, Zhao and Zhang}]{ref1}
\bibinfo{author}{Hu, L.}, \bibinfo{author}{Yang, D.}, \bibinfo{author}{Zhao,
  D.}, \bibinfo{author}{Zhang, J.}, \bibinfo{year}{2024}.
\newblock \bibinfo{title}{Infrared small target detection method based on
  improved non-convex estimation and asymmetric spatial-temporal
  regulariaztion}.
\newblock \bibinfo{journal}{Journal of National University of Defense
  Technonlogy} \bibinfo{volume}{46}, \bibinfo{pages}{180--194}.
%Type = Article
\bibitem[{Huang et~al.(2019)Huang, Liu, He, Zhang and Peng}]{ref12}
\bibinfo{author}{Huang, S.}, \bibinfo{author}{Liu, Y.}, \bibinfo{author}{He,
  Y.}, \bibinfo{author}{Zhang, T.}, \bibinfo{author}{Peng, Z.},
  \bibinfo{year}{2019}.
\newblock \bibinfo{title}{Structure-adaptive clutter suppression for infrared
  small target detection: Chain-growth filtering}.
\newblock \bibinfo{journal}{Remote Sensing} \bibinfo{volume}{12},
  \bibinfo{pages}{47}.
\newblock \DOIprefix\doi{10.1109/TAES.2019.2946678}.
%Type = Article
\bibitem[{Jia et~al.(2024)Jia, Cheng and Chen}]{ref38}
\bibinfo{author}{Jia, G.}, \bibinfo{author}{Cheng, Y.}, \bibinfo{author}{Chen,
  T.}, \bibinfo{year}{2024}.
\newblock \bibinfo{title}{Irgraphseg: infrared small target detection based on
  hierarchical gnn}.
\newblock \bibinfo{journal}{IEEE Geoscience and Remote Sensing Letters}
  \bibinfo{volume}{21}, \bibinfo{pages}{1--5}.
\newblock \DOIprefix\doi{10.1109/LGRS.2024.3374431}.
%Type = Article
\bibitem[{Kou et~al.(2023)Kou, Wang, Peng, Zhao, Chen, Han, Huang, Yu and
  Fu}]{ref3}
\bibinfo{author}{Kou, R.}, \bibinfo{author}{Wang, C.}, \bibinfo{author}{Peng,
  Z.}, \bibinfo{author}{Zhao, Z.}, \bibinfo{author}{Chen, Y.},
  \bibinfo{author}{Han, J.}, \bibinfo{author}{Huang, F.}, \bibinfo{author}{Yu,
  Y.}, \bibinfo{author}{Fu, Q.}, \bibinfo{year}{2023}.
\newblock \bibinfo{title}{Infrared small target segmentation networks: A
  survey}.
\newblock \bibinfo{journal}{Pattern Recognition} \bibinfo{volume}{143},
  \bibinfo{pages}{109788}.
\newblock \DOIprefix\doi{10.1016/j.patcog.2023.109788}.
%Type = Article
\bibitem[{Li et~al.(2022)Li, Xiao, Wang, Wang, Lin, Li, An and Guo}]{ref30}
\bibinfo{author}{Li, B.}, \bibinfo{author}{Xiao, C.}, \bibinfo{author}{Wang,
  L.}, \bibinfo{author}{Wang, Y.}, \bibinfo{author}{Lin, Z.},
  \bibinfo{author}{Li, M.}, \bibinfo{author}{An, W.}, \bibinfo{author}{Guo,
  Y.}, \bibinfo{year}{2022}.
\newblock \bibinfo{title}{Dense nested attention network for infrared small
  target detection}.
\newblock \bibinfo{journal}{IEEE Transactions on Image Processing}
  \bibinfo{volume}{32}, \bibinfo{pages}{1745--1758}.
\newblock \DOIprefix\doi{10.1109/TIP.2022.3199107}.
%Type = Inproceedings
\bibitem[{Li et~al.(2019)Li, Wang, Hu and Yang}]{ref42}
\bibinfo{author}{Li, X.}, \bibinfo{author}{Wang, W.}, \bibinfo{author}{Hu, X.},
  \bibinfo{author}{Yang, J.}, \bibinfo{year}{2019}.
\newblock \bibinfo{title}{Selective kernel networks}, in:
  \bibinfo{booktitle}{Proceedings of the IEEE/CVF conference on computer vision
  and pattern recognition}, pp. \bibinfo{pages}{510--519}.
\newblock \DOIprefix\doi{10.1109/CVPR.2019.00060}.
%Type = Article
\bibitem[{Liu et~al.(2023a)Liu, Gao, Chen, Meng, Zuo and Gao}]{ref35}
\bibinfo{author}{Liu, F.}, \bibinfo{author}{Gao, C.}, \bibinfo{author}{Chen,
  F.}, \bibinfo{author}{Meng, D.}, \bibinfo{author}{Zuo, W.},
  \bibinfo{author}{Gao, X.}, \bibinfo{year}{2023}a.
\newblock \bibinfo{title}{Infrared small and dim target detection with
  transformer under complex backgrounds}.
\newblock \bibinfo{journal}{IEEE Transactions on Image Processing}
  \bibinfo{volume}{32}, \bibinfo{pages}{5921--5932}.
\newblock \DOIprefix\doi{10.1109/TIP.2023.3326396}.
%Type = Inproceedings
\bibitem[{Liu et~al.(2024)Liu, Liu, Zheng, Wang and Fu}]{ref34}
\bibinfo{author}{Liu, Q.}, \bibinfo{author}{Liu, R.}, \bibinfo{author}{Zheng,
  B.}, \bibinfo{author}{Wang, H.}, \bibinfo{author}{Fu, Y.},
  \bibinfo{year}{2024}.
\newblock \bibinfo{title}{Infrared small target detection with scale and
  location sensitivity}, in: \bibinfo{booktitle}{Proceedings of the IEEE/CVF
  Conference on Computer Vision and Pattern Recognition}, pp.
  \bibinfo{pages}{17490--17499}.
%Type = Inproceedings
\bibitem[{Liu et~al.(2016)Liu, Anguelov, Erhan, Szegedy, Reed, Fu and
  Berg}]{ref55}
\bibinfo{author}{Liu, W.}, \bibinfo{author}{Anguelov, D.},
  \bibinfo{author}{Erhan, D.}, \bibinfo{author}{Szegedy, C.},
  \bibinfo{author}{Reed, S.}, \bibinfo{author}{Fu, C.Y.},
  \bibinfo{author}{Berg, A.C.}, \bibinfo{year}{2016}.
\newblock \bibinfo{title}{Ssd: Single shot multibox detector}, in:
  \bibinfo{booktitle}{European conference on computer vision},
  \bibinfo{organization}{Springer}. pp. \bibinfo{pages}{21--37}.
%Type = Inproceedings
\bibitem[{Liu et~al.(2023b)Liu, Lu, Fu and Cao}]{ref43}
\bibinfo{author}{Liu, W.}, \bibinfo{author}{Lu, H.}, \bibinfo{author}{Fu, H.},
  \bibinfo{author}{Cao, Z.}, \bibinfo{year}{2023}b.
\newblock \bibinfo{title}{Learning to upsample by learning to sample}, in:
  \bibinfo{booktitle}{Proceedings of the IEEE/CVF International Conference on
  Computer Vision}, pp. \bibinfo{pages}{6004--6014}.
\newblock \DOIprefix\doi{10.1109/ICCV51070.2023.00554}.
%Type = Article
\bibitem[{Liu and Peng(2022)}]{ref17}
\bibinfo{author}{Liu, Y.}, \bibinfo{author}{Peng, Z.}, \bibinfo{year}{2022}.
\newblock \bibinfo{title}{Infrared small target detection based on
  resampling-guided image model}.
\newblock \bibinfo{journal}{IEEE Geoscience and Remote Sensing Letters}
  \bibinfo{volume}{19}, \bibinfo{pages}{1--5}.
\newblock \DOIprefix\doi{10.1109/LGRS.2021.3087799}.
%Type = Article
\bibitem[{Luo et~al.(2010)Luo, Wang, Chen and Yu}]{ref0}
\bibinfo{author}{Luo, H.}, \bibinfo{author}{Wang, F.}, \bibinfo{author}{Chen,
  Z.}, \bibinfo{author}{Yu, L.}, \bibinfo{year}{2010}.
\newblock \bibinfo{title}{Infrared target detecting based on symmetrical
  displaced frame difference and optical flow estimation}.
\newblock \bibinfo{journal}{Acta Optica Sinica} \bibinfo{volume}{30},
  \bibinfo{pages}{1715--1720}.
%Type = Article
\bibitem[{M(2011)}]{ref50}
\bibinfo{author}{M, C.}, \bibinfo{year}{2011}.
\newblock \bibinfo{title}{Visual attention: the past 25 years}.
\newblock \bibinfo{journal}{Vision Research: An International Journal in Visual
  Science} \bibinfo{volume}{51}, \bibinfo{pages}{1484--1525}.
\newblock \DOIprefix\doi{10.1016/j.visres.2011.04.012}.
%Type = Article
\bibitem[{Nian et~al.(2023)Nian, Jiang, Shi and Zhang}]{ref45}
\bibinfo{author}{Nian, B.}, \bibinfo{author}{Jiang, B.}, \bibinfo{author}{Shi,
  H.}, \bibinfo{author}{Zhang, Y.}, \bibinfo{year}{2023}.
\newblock \bibinfo{title}{Local contrast attention guide network for detecting
  infrared small targets}.
\newblock \bibinfo{journal}{IEEE Transactions on Geoscience and Remote Sensing}
  \bibinfo{volume}{61}, \bibinfo{pages}{1}.
\newblock \DOIprefix\doi{10.1109/TGRS.2023.3266447}.
%Type = Inproceedings
\bibitem[{Rahman and Wang(2016)}]{ref52}
\bibinfo{author}{Rahman, M.A.}, \bibinfo{author}{Wang, Y.},
  \bibinfo{year}{2016}.
\newblock \bibinfo{title}{Optimizing intersection-over-union in deep neural
  networks for image segmentation}, in: \bibinfo{booktitle}{International
  symposium on visual computing}, \bibinfo{organization}{Springer}. pp.
  \bibinfo{pages}{234--244}.
%Type = Inproceedings
\bibitem[{Redmon et~al.(2016)Redmon, Divvala, Girshick and Farhadi}]{ref54}
\bibinfo{author}{Redmon, J.}, \bibinfo{author}{Divvala, S.},
  \bibinfo{author}{Girshick, R.}, \bibinfo{author}{Farhadi, A.},
  \bibinfo{year}{2016}.
\newblock \bibinfo{title}{You only look once: Unified, real-time object
  detection}, in: \bibinfo{booktitle}{Proceedings of the IEEE/CVF conference on
  computer vision and pattern recognition}.
%Type = Article
\bibitem[{Shi et~al.(2024)Shi, Lin, Wei, Xian, Chen and Lin}]{ref37}
\bibinfo{author}{Shi, Y.}, \bibinfo{author}{Lin, Y.}, \bibinfo{author}{Wei,
  P.}, \bibinfo{author}{Xian, X.}, \bibinfo{author}{Chen, T.},
  \bibinfo{author}{Lin, L.}, \bibinfo{year}{2024}.
\newblock \bibinfo{title}{Diff-mosaic: augmenting realistic representations in
  infrared small target detection via diffusion prior}.
\newblock \bibinfo{journal}{IEEE Transactions on Geoscience and Remote Sensing}
  \bibinfo{volume}{61}, \bibinfo{pages}{1--11}.
\newblock \DOIprefix\doi{10.1109/TGRS.2024.3408045}.
%Type = Article
\bibitem[{Sun et~al.(2023)Sun, Bai, Yang and Bai}]{ref5}
\bibinfo{author}{Sun, H.}, \bibinfo{author}{Bai, J.}, \bibinfo{author}{Yang,
  F.}, \bibinfo{author}{Bai, X.}, \bibinfo{year}{2023}.
\newblock \bibinfo{title}{Receptive-field and direction induced attention
  network for infrared dim small target detection with a large-scale dataset
  irdst}.
\newblock \bibinfo{journal}{IEEE Transactions on Geoscience and Remote Sensing}
  \bibinfo{volume}{61}, \bibinfo{pages}{1--13}.
\newblock \DOIprefix\doi{10.1109/TGRS.2023.3235150}.
%Type = Article
\bibitem[{Tianxiang et~al.(2024)Tianxiang, Zhentao, Tao, Qi, Yue, Bin, Jieping
  and Nenghai}]{ref56}
\bibinfo{author}{Tianxiang, C.}, \bibinfo{author}{Zhentao, T.},
  \bibinfo{author}{Tao, G.}, \bibinfo{author}{Qi, C.}, \bibinfo{author}{Yue,
  W.}, \bibinfo{author}{Bin, L.}, \bibinfo{author}{Jieping, Y.},
  \bibinfo{author}{Nenghai, Y.}, \bibinfo{year}{2024}.
\newblock \bibinfo{title}{Mim-istd: Mamba-in-mamba for efficient infrared small
  target detection}.
\newblock \bibinfo{journal}{arXiv preprint arXiv:2403.02148} .
%Type = Article
\bibitem[{Wang et~al.(2024)Wang, Song and Dong}]{ref4}
\bibinfo{author}{Wang, B.}, \bibinfo{author}{Song, Y.}, \bibinfo{author}{Dong,
  X.}, \bibinfo{year}{2024}.
\newblock \bibinfo{title}{Indistinguishable points attention-aware network for
  infrared small object detection}.
\newblock \bibinfo{journal}{Chinese Optics} \bibinfo{volume}{17},
  \bibinfo{pages}{538--547}.
%Type = Article
\bibitem[{Wang et~al.(2021)Wang, Tao, Kong and Peng}]{ref19}
\bibinfo{author}{Wang, G.}, \bibinfo{author}{Tao, B.}, \bibinfo{author}{Kong,
  X.}, \bibinfo{author}{Peng, Z.}, \bibinfo{year}{2021}.
\newblock \bibinfo{title}{Infrared small target detection using nonoverlapping
  patch spatial--temporal tensor factorization with capped nuclear norm
  regularization}.
\newblock \bibinfo{journal}{IEEE Transactions on Geoscience and Remote Sensing}
  \bibinfo{volume}{60}, \bibinfo{pages}{1--17}.
\newblock \DOIprefix\doi{10.1109/TGRS.2021.3126608}.
%Type = Inproceedings
\bibitem[{Wang et~al.(2019)Wang, Zhou and Wang}]{ref28}
\bibinfo{author}{Wang, H.}, \bibinfo{author}{Zhou, L.}, \bibinfo{author}{Wang,
  L.}, \bibinfo{year}{2019}.
\newblock \bibinfo{title}{Miss detection vs. false alarm: Adversarial learning
  for small object segmentation in infrared images}, in:
  \bibinfo{booktitle}{Proceedings of the IEEE/CVF International Conference on
  Computer Vision}, pp. \bibinfo{pages}{8509--8518}.
\newblock \DOIprefix\doi{10.1109/ICCV.2019.00860}.
%Type = Article
\bibitem[{Wang et~al.(2022)Wang, Du, Liu and Cao}]{ref48}
\bibinfo{author}{Wang, K.}, \bibinfo{author}{Du, S.}, \bibinfo{author}{Liu,
  C.}, \bibinfo{author}{Cao, Z.}, \bibinfo{year}{2022}.
\newblock \bibinfo{title}{Interior attention-aware network for infrared small
  target detection}.
\newblock \bibinfo{journal}{IEEE Transactions on Geoscience and Remote Sensing}
  \bibinfo{volume}{60}, \bibinfo{pages}{1--13}.
\newblock \DOIprefix\doi{10.1109/TGRS.2022.3163410}.
%Type = Article
\bibitem[{Wei et~al.(2016)Wei, You and Li}]{ref24}
\bibinfo{author}{Wei, Y.}, \bibinfo{author}{You, X.}, \bibinfo{author}{Li, H.},
  \bibinfo{year}{2016}.
\newblock \bibinfo{title}{Multiscale patch-based contrast measure for small
  infrared target detection}.
\newblock \bibinfo{journal}{Pattern Recognition} \bibinfo{volume}{58},
  \bibinfo{pages}{216--226}.
\newblock \DOIprefix\doi{10.1016/j.patcog.2016.04.002}.
%Type = Article
\bibitem[{Wu et~al.(2024)Wu, Fan, Min, Qin and Yu}]{ref21}
\bibinfo{author}{Wu, A.}, \bibinfo{author}{Fan, X.}, \bibinfo{author}{Min, L.},
  \bibinfo{author}{Qin, W.}, \bibinfo{author}{Yu, L.}, \bibinfo{year}{2024}.
\newblock \bibinfo{title}{Dim and small target detection based on local feature
  prior and tensor train nuclear norm}.
\newblock \bibinfo{journal}{IEEE Photonics Journal} \bibinfo{volume}{16}.
\newblock \DOIprefix\doi{10.1109/JPHOT.2024.3351189}.
%Type = Article
\bibitem[{Wu et~al.(2023)Wu, Li, Luo, Wang, Xiao, Liu, Yang, An and
  Guo}]{ref36}
\bibinfo{author}{Wu, T.}, \bibinfo{author}{Li, B.}, \bibinfo{author}{Luo, Y.},
  \bibinfo{author}{Wang, Y.}, \bibinfo{author}{Xiao, C.}, \bibinfo{author}{Liu,
  T.}, \bibinfo{author}{Yang, J.}, \bibinfo{author}{An, W.},
  \bibinfo{author}{Guo, Y.}, \bibinfo{year}{2023}.
\newblock \bibinfo{title}{Mtu-net: multilevel transunet for space-based
  infrared tiny ship detection}.
\newblock \bibinfo{journal}{IEEE Transactions on Geoscience and Remote Sensing}
  \bibinfo{volume}{61}, \bibinfo{pages}{1--15}.
\newblock \DOIprefix\doi{10.1109/TGRS.2023.3235002}.
%Type = Article
\bibitem[{Wu et~al.(2022)Wu, Hong and Chanussot}]{ref33}
\bibinfo{author}{Wu, X.}, \bibinfo{author}{Hong, D.},
  \bibinfo{author}{Chanussot, J.}, \bibinfo{year}{2022}.
\newblock \bibinfo{title}{Uiu-net: U-net in u-net for infrared small object
  detection}.
\newblock \bibinfo{journal}{IEEE Transactions on Image Processing}
  \bibinfo{volume}{32}, \bibinfo{pages}{364--376}.
\newblock \DOIprefix\doi{10.1109/TIP.2022.3228497}.
%Type = Article
\bibitem[{Xia et~al.(2024)Xia, Chen, Huang, Hu and Chen}]{ref20}
\bibinfo{author}{Xia, C.}, \bibinfo{author}{Chen, S.}, \bibinfo{author}{Huang,
  R.}, \bibinfo{author}{Hu, J.}, \bibinfo{author}{Chen, Z.},
  \bibinfo{year}{2024}.
\newblock \bibinfo{title}{Separable spatial-temporal patch-tensor pair
  completion for infrared small target detection}.
\newblock \bibinfo{journal}{IEEE Transactions on Geoscience and Remote Sensing}
  \bibinfo{volume}{62}.
\newblock \DOIprefix\doi{10.1109/TGRS.2024.3358831}.
%Type = Article
\bibitem[{Xia et~al.(2019)Xia, Li, Zhao and Shu}]{ref25}
\bibinfo{author}{Xia, C.}, \bibinfo{author}{Li, X.}, \bibinfo{author}{Zhao,
  L.}, \bibinfo{author}{Shu, R.}, \bibinfo{year}{2019}.
\newblock \bibinfo{title}{Infrared small target detection based on multiscale
  local contrast measure using local energy factor}.
\newblock \bibinfo{journal}{IEEE Geoscience and Remote Sensing Letters}
  \bibinfo{volume}{17}, \bibinfo{pages}{157--161}.
\newblock \DOIprefix\doi{10.1109/LGRS.2019.2914432}.
%Type = Article
\bibitem[{Yang et~al.(2024)Yang, Mu, Dong, Zhang, Wang, Ke, Yang and
  He}]{ref49}
\bibinfo{author}{Yang, H.}, \bibinfo{author}{Mu, T.}, \bibinfo{author}{Dong,
  Z.}, \bibinfo{author}{Zhang, Z.}, \bibinfo{author}{Wang, B.},
  \bibinfo{author}{Ke, W.}, \bibinfo{author}{Yang, Q.}, \bibinfo{author}{He,
  Z.}, \bibinfo{year}{2024}.
\newblock \bibinfo{title}{Pbt: Progressive background-aware transformer for
  infrared small target detection}.
\newblock \bibinfo{journal}{IEEE Transactions on Geoscience and Remote Sensing}
  \DOIprefix\doi{10.1109/TGRS.2024.3415080}.
%Type = Article
\bibitem[{Yu et~al.(2022)Yu, Liu, Wu, Hu, Xia, Lan and Liu}]{ref39}
\bibinfo{author}{Yu, C.}, \bibinfo{author}{Liu, Y.}, \bibinfo{author}{Wu, S.},
  \bibinfo{author}{Hu, Z.}, \bibinfo{author}{Xia, X.}, \bibinfo{author}{Lan,
  D.}, \bibinfo{author}{Liu, X.}, \bibinfo{year}{2022}.
\newblock \bibinfo{title}{Infrared small target detection based on multiscale
  local contrast learning networks}.
\newblock \bibinfo{journal}{Infrared Physics \& Technology}
  \bibinfo{volume}{123}, \bibinfo{pages}{104107}.
\newblock \DOIprefix\doi{10.1016/j.infrared.2022.104107}.
%Type = Article
\bibitem[{Yu and Koltun(2015)}]{ref44}
\bibinfo{author}{Yu, F.}, \bibinfo{author}{Koltun, V.}, \bibinfo{year}{2015}.
\newblock \bibinfo{title}{Multi-scale context aggregation by dilated
  convolutions}.
\newblock \bibinfo{journal}{arXiv preprint arXiv:1511.07122} .
%Type = Article
\bibitem[{Yuan et~al.(2024)Yuan, Qin, Yan, Akhtar and Mian}]{ref6}
\bibinfo{author}{Yuan, S.}, \bibinfo{author}{Qin, H.}, \bibinfo{author}{Yan,
  X.}, \bibinfo{author}{Akhtar, N.}, \bibinfo{author}{Mian, A.},
  \bibinfo{year}{2024}.
\newblock \bibinfo{title}{Sctransnet: Spatial-channel cross transformer network
  for infrared small target detection}.
\newblock \bibinfo{journal}{IEEE Transactions on Geoscience and Remote Sensing}
  \bibinfo{volume}{62}, \bibinfo{pages}{1--15}.
\newblock \DOIprefix\doi{10.1109/TGRS.2024.3383649}.
%Type = Article
\bibitem[{Zeng et~al.(2006)Zeng, Li and Peng}]{ref9}
\bibinfo{author}{Zeng, M.}, \bibinfo{author}{Li, J.}, \bibinfo{author}{Peng,
  Z.}, \bibinfo{year}{2006}.
\newblock \bibinfo{title}{The design of top-hat morphological filter and
  application to infrared target detection}.
\newblock \bibinfo{journal}{Infrared physics \& technology}
  \bibinfo{volume}{48}, \bibinfo{pages}{67--76}.
\newblock \DOIprefix\doi{10.1016/j.infrared.2005.04.006}.
%Type = Inproceedings
\bibitem[{Zhang et~al.(2022a)Zhang, Wu, Zhang, Zhu, Lin, Zhang, Sun, He,
  Mueller, Manmatha et~al.}]{ref40}
\bibinfo{author}{Zhang, H.}, \bibinfo{author}{Wu, C.}, \bibinfo{author}{Zhang,
  Z.}, \bibinfo{author}{Zhu, Y.}, \bibinfo{author}{Lin, H.},
  \bibinfo{author}{Zhang, Z.}, \bibinfo{author}{Sun, Y.}, \bibinfo{author}{He,
  T.}, \bibinfo{author}{Mueller, J.}, \bibinfo{author}{Manmatha, R.}, et~al.,
  \bibinfo{year}{2022}a.
\newblock \bibinfo{title}{Resnest: split-attention networks}, in:
  \bibinfo{booktitle}{Proceedings of the IEEE/CVF conference on computer vision
  and pattern recognition}, pp. \bibinfo{pages}{2736--2746}.
\newblock \DOIprefix\doi{10.1109/CVPRW56347.2022.00309}.
%Type = Article
\bibitem[{Zhang et~al.(2024)Zhang, Yang, Zheng, Zhang and Zhang}]{ref27}
\bibinfo{author}{Zhang, L.}, \bibinfo{author}{Yang, H.},
  \bibinfo{author}{Zheng, Q.}, \bibinfo{author}{Zhang, Y.},
  \bibinfo{author}{Zhang, D.}, \bibinfo{year}{2024}.
\newblock \bibinfo{title}{Infrared small target detection based on density peak
  search and local features}.
\newblock \bibinfo{journal}{IET Signal Processing} \bibinfo{volume}{2024},
  \bibinfo{pages}{6814362}.
\newblock \DOIprefix\doi{10.1016/j.infrared.2024.105315}.
%Type = Inproceedings
\bibitem[{Zhang et~al.(2022b)Zhang, Zhang, Yang, Bai, Zhang and Guo}]{ref32}
\bibinfo{author}{Zhang, M.}, \bibinfo{author}{Zhang, R.},
  \bibinfo{author}{Yang, Y.}, \bibinfo{author}{Bai, H.},
  \bibinfo{author}{Zhang, J.}, \bibinfo{author}{Guo, J.},
  \bibinfo{year}{2022}b.
\newblock \bibinfo{title}{Isnet: shape matters for infrared small target
  detection}, in: \bibinfo{booktitle}{Proceedings of the IEEE/CVF Conference on
  Computer Vision and Pattern Recognition}, pp. \bibinfo{pages}{877--886}.
\newblock \DOIprefix\doi{10.1109/CVPR52688.2022.00095}.
%Type = Article
\bibitem[{Zhang et~al.(2023)Zhang, Gao, Liu, Hussain, Waqas, Halim and
  Li}]{ref46}
\bibinfo{author}{Zhang, T.}, \bibinfo{author}{Gao, Z.}, \bibinfo{author}{Liu,
  Z.}, \bibinfo{author}{Hussain, S.F.}, \bibinfo{author}{Waqas, M.},
  \bibinfo{author}{Halim, Z.}, \bibinfo{author}{Li, Y.}, \bibinfo{year}{2023}.
\newblock \bibinfo{title}{Infrared ship target segmentation based on
  adversarial domain adaptation}.
\newblock \bibinfo{journal}{Knowledge-Based Systems} \bibinfo{volume}{265},
  \bibinfo{pages}{110344}.
\newblock \DOIprefix\doi{10.1016/j.knosys.2023.110344}.
%Type = Article
\bibitem[{Zhao et~al.(2024)Zhao, Shi, Yu and Liu}]{ref8}
\bibinfo{author}{Zhao, J.}, \bibinfo{author}{Shi, Z.}, \bibinfo{author}{Yu,
  C.}, \bibinfo{author}{Liu, Y.}, \bibinfo{year}{2024}.
\newblock \bibinfo{title}{Multi-scale direction-aware network for infrared
  small target detection}.
\newblock \bibinfo{journal}{arXiv preprint arXiv:2406.02037} .
%Type = Article
\bibitem[{Zhao et~al.(2020)Zhao, Li, Li, Tao, Li and Zhang}]{ref13}
\bibinfo{author}{Zhao, M.}, \bibinfo{author}{Li, L.}, \bibinfo{author}{Li, W.},
  \bibinfo{author}{Tao, R.}, \bibinfo{author}{Li, L.}, \bibinfo{author}{Zhang,
  W.}, \bibinfo{year}{2020}.
\newblock \bibinfo{title}{Infrared small-target detection based on multiple
  morphological profiles}.
\newblock \bibinfo{journal}{IEEE Transactions on Geoscience and Remote Sensing}
  \bibinfo{volume}{59}, \bibinfo{pages}{6077--6091}.
\newblock \DOIprefix\doi{10.1109/TGRS.2020.3022863}.
%Type = Article
\bibitem[{Zhao et~al.(2022)Zhao, Li, Li, Hu, Ma and Tao}]{ref23}
\bibinfo{author}{Zhao, M.}, \bibinfo{author}{Li, W.}, \bibinfo{author}{Li, L.},
  \bibinfo{author}{Hu, J.}, \bibinfo{author}{Ma, P.}, \bibinfo{author}{Tao,
  R.}, \bibinfo{year}{2022}.
\newblock \bibinfo{title}{Single-frame infrared small-target detection: A
  survey}.
\newblock \bibinfo{journal}{IEEE Geoscience and Remote Sensing Magazine}
  \bibinfo{volume}{10}, \bibinfo{pages}{87--119}.
\newblock \DOIprefix\doi{10.1109/MGRS.2022.3145502}.

\end{thebibliography}


\end{document}